\documentclass[sigconf]{acmart}

\usepackage{multirow}
\usepackage{booktabs}

\AtBeginDocument{%
	}

\setcopyright{acmcopyright}
\copyrightyear{2023}
\acmYear{2023}
\acmDOI{XXXXXXX.XXXXXXX}

\acmConference[MM ’23]{Make sure to enter the correct  conference title from your rights confirmation emai}{October 29 - November 3, 2023}{Ottawa, Canada}
\acmPrice{15.00}
\acmISBN{978-1-4503-XXXX-X/18/06}

\acmSubmissionID{1092}

\renewcommand\footnotetextcopyrightpermission[1]{}
\begin{document}
	
	\title{Masked Contrastive Graph Representation Learning for Age Estimation}

 	\author{Yuntao Shou}
	\affiliation{
		  \institution{Xi'an Jiaotong University}
		  \city{Xi'an}
		  \country{China}}
	\email{sytzyq@gmail.com}
	
	\author{Xiangyong Cao}
	\authornote{Corresponding Author}
	\affiliation{
	\institution{Xi'an Jiaotong University}
	\city{Xi'an}
	\country{China}}
	\email{caoxiangyong@mail.xjtu.edu.cn}

	\author{Deyu Meng}
	\affiliation{
	\institution{Xi'an Jiaotong University}
	\city{Xi'an}
	\country{China}}
	\email{dymeng@mail.xjtu.edu.cn}

	\renewcommand{\shortauthors}{Trovato et al.}
	
\begin{abstract}
  Age estimation of face images is a crucial task with various practical applications in areas such as video surveillance and Internet access control. While deep learning-based age estimation frameworks, e.g., convolutional neural network (CNN), multi-layer perceptrons (MLP), and transformers have shown remarkable performance, they have limitations when modelling complex or irregular objects in an image that contains a large amount of redundant information. To address this issue, this paper utilizes the robustness property of graph representation learning in dealing with image redundancy information and proposes a novel Masked Contrastive Graph Representation Learning (MCGRL) method for age estimation. Specifically, our approach first leverages CNN to extract semantic features of the image, which are then partitioned into patches that serve as nodes in the graph. Then, we use a masked graph convolutional network (GCN) to derive image-based node representations that capture rich structural information. Finally, we incorporate multiple losses to explore the complementary relationship between structural information and semantic features, which improves the feature representation capability of GCN. Experimental results on real-world face image datasets demonstrate the superiority of our proposed method over other state-of-the-art age estimation approaches.
	\end{abstract}

	
	
	
	\maketitle
	
	\section{Introduction}
	\begin{figure}
		\centering
		\includegraphics[width=1\linewidth]{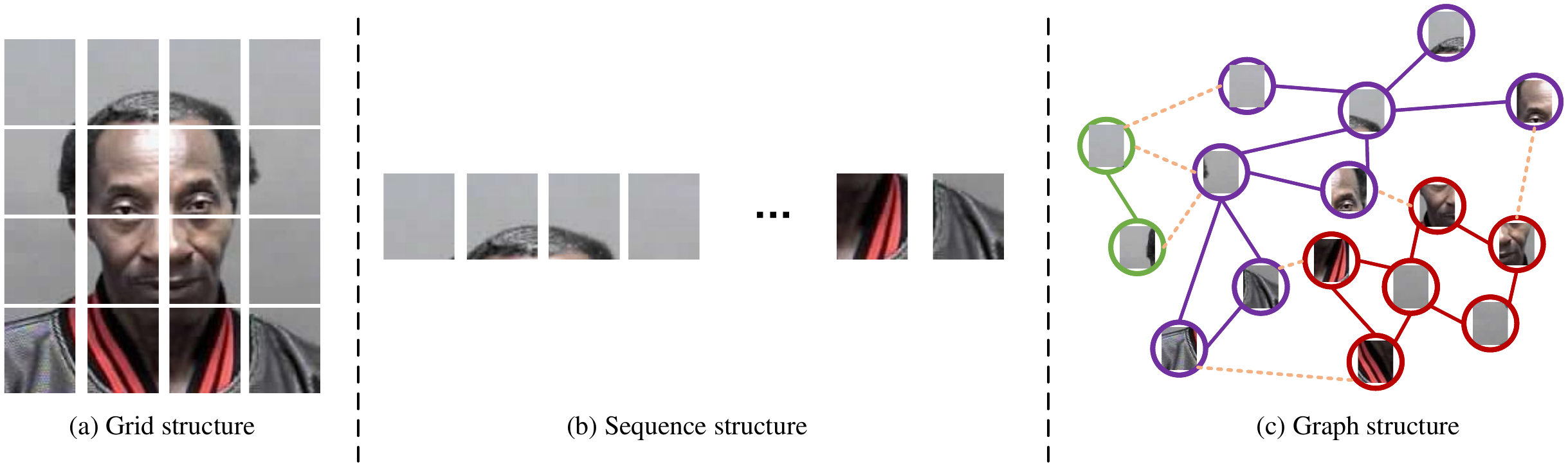}
		\caption{Illustrative examples of grid, sequence, and graph-structured representations of an image. (a) In a grid structure, pixels or patches are ordered according to a regular structure (e.g., square or rectangle). (b) In the sequential structure, images are arranged as a sequence of patches. (c) In the graph structure, nodes are connected by adjacent patches and can model irregular objects.}
		\label{fig:}
	\end{figure}
	
	Age estimation of face images has garnered significant interest from researchers in recent years, owing to the growing prevalence of deep learning techniques in the field of computer vision. The task of age estimation involves predicting a person's age based on the semantic features present in their facial image, and it is widely used in video surveillance and minors' anti-addiction systems. For example, in the anti-addiction system for minors, it is possible to prevent minors from indulging in games by estimating the age of players.
	
	\begin{figure}
		\centering
		\includegraphics[width=1\linewidth]{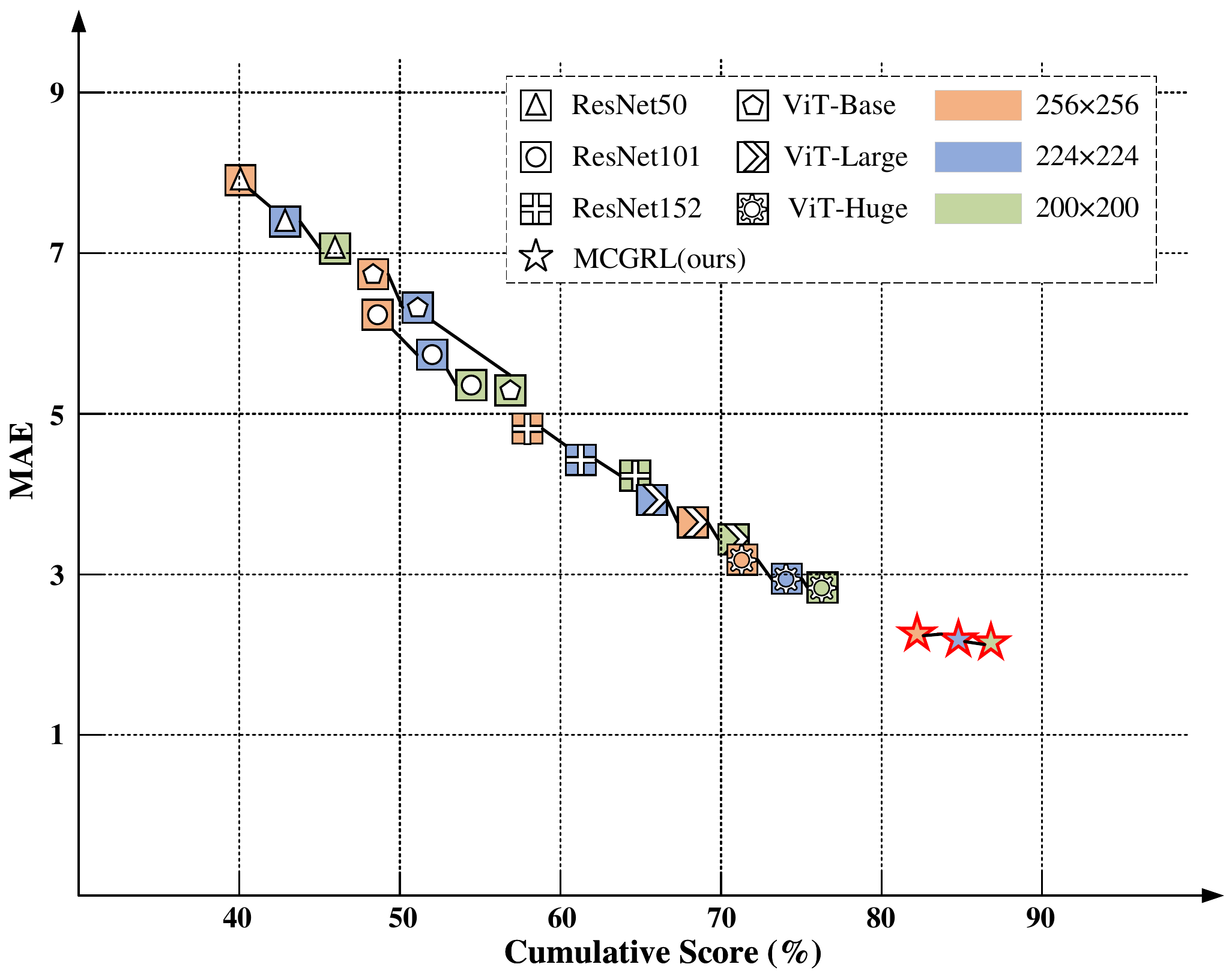}
		\caption{The variations of MAE and a cumulative score of our method MCGRL and previous CNN and ViT methods with different image sizes on the MORPH dataset.}
		\label{fig:intro2}
	\end{figure}
	
	In the field of image processing, CNN and Transformer have become common network architectures, which play a decisive role in feature extraction capabilities. Different network architectures can model images in different ways. As shown in Fig. 1(a), CNNs usually process image data in a regular grid manner. As shown in Fig. 1(b), the Transformer treats image data as a sequence of pixels or patches. Both CNN and Transformer process images in Euclidean space and cannot be applied to data in non-Euclidean space. This modelling method is inflexible for complex and irregular objects. As shown in Fig. 1(c), the graph-based modelling approach can flexibly model complex and irregular objects. For example, a human face is composed of hair, eyes, nose, mouth, and ears, and these parts can naturally form a graph structure. By extracting the information in the graph structure, we can predict the age of the person. Furthermore, regular grid structures and sequence structures can be viewed as a special case of graph structures. However, the modelling method based on CNN and Transformer is easy to introduce redundant information when dealing with irregular data, which will bring noise interference to the model prediction age. As shown in Figure 2, we cut the areas that are not related to the face in the image and then input them into CNN or Vision Transformer (ViT) for age estimation, their MAE value will decrease and the cumulative score will increase. While using graph structure for age estimation, the performance of the model remains almost the same regardless of whether the model is clipped to regions not related to the face. In summary, the graph structure-based modelling method is more robust in dealing with images with a large amount of reductant information.
	
	Therefore, this paper attempts to utilize the graph representation learning technique to deal with the image redundancy information issue for the age estimation task. Specifically, we propose a novel method called Masked Contrastive Graph Representation Learning for Age Estimation (MCGRL). This method is capable of modelling complex and irregular objects while more efficiently fusing complementary semantic information between structural information and semantic features. Firstly, we extract semantic features in images using CNN as anchor embeddings and divide images into patches as nodes in the graph. Secondly, to improve node feature representation, we utilize a masked graph convolutional neural network (GCN) to aggregate node information after masking some nodes in the graph. Thirdly, we generate positive embedding samples using masked GCN and neighbour sampling, and negative embedding samples using random row shuffling. Finally, we employ multiple loss functions to minimize the spatial distance between the anchor and positive embeddings, while maximizing the distance between positive and negative samples. Therefore, this fusion of complementary semantic information between structural information and semantic features can be achieved by our proposed MCGRL method.
	
	
	In summary, the contributions of our method are as follows:
	
	\begin{itemize}
		\item This paper proposes a novel Masked Contrastive Graph Representation Learning (MCGRL) method to estimate the age of face images. The MCGRL method can alleviate the inflexibility issue of the existing age estimation methods in modeling irregular objects.	
		\item The MCGRL method provides a general framework to model irregular objects in the image by using a graph structure. The representation ability of node semantic information can be improved by a graph convolutional network with a mask. Furthermore, the complementary semantic information between structural information and semantic features is captured by contrastive learning methodology.
		\item Extensive experiments demonstrate the superiority of the proposed MCGRL method compared with other state-of-the-art age estimation methods. 
	\end{itemize}
	
	\section{Related Work}
	
	\subsection{Age Estimation Methods}
	The age of face images is estimated to be widely used in many areas in the real world, and it has great application value. The age estimation method can be roughly divided into two categories, i.e., machine learning methods and deep learning methods.
	
	\textbf{Machine learning methods:} The main idea of machine learning methods is to use some traditional regression methods to learn from the data extracted by hand, and obtain a coherent curve with the good generalization ability. Typical methods include Moving Window Regression (MWR) \cite{shin2022moving}, Consistent Ordinal Regression (CORAL) \cite{cao2020rank}, Quantifying Facial Age by Posterior \cite{zhang2017quantifying}, Ranking SVM (RS) \cite{cao2012human}, etc \cite{li2019bridgenet, shen2018deep}.
	
	\textbf{Deep learning methods:} The main idea of deep learning methods is to extract deep image features in an end-to-end manner from the primitive face image and then combine these image feature information to perform age estimation tasks. Currently, deep learning-based age estimation methods include CNN \cite{ levi2015age}, attention network \cite{wang2022improving}, and hybrid neural networks \cite{xie2015hybrid}. The CNN architecture is mainly to extract the deep-seated features of the local area in the image through convolution operations, and the local characteristics are combined through the full connection layer to make age prediction, e.g., Cascade Context-Based Age Estimation (C3AE) \cite{zhang2019c3ae}, Agenet \cite{liu2015agenet}, etc \cite{shoba2022adaptive}. The attention mechanism mainly uses global modelling capabilities to pay attention to the key areas in the image and combine key information to perform age prediction, e.g., Attention-Based Dynamic Patch Fusion (ADPF) \cite{wang2022improving}, Hierarchical Attention-Based Age Estimation \cite{hiba2021hierarchical}, etc \cite{zhang2019fine}, \cite{shi2020fetal}. The hybrid neural network predicts age prediction by combining the characteristic extraction ability of deep learning and the fitting ability of traditional machine learning, e.g., \cite{xie2015hybrid}, \cite{tang2014compressed}.
	
	\subsection{Graph Neural Network}
	The most commonly used network architectures in the field of computer vision are Convolutional Neural Networks (CNN) \cite{liu2022convnet, huang2018condensenet, allen2019can} and Vision Transformers (ViT) \cite{han2022survey, arnab2021vivit, mao2022towards, liu2021swin}. CNN and ViT can extract deep features for regular images, but have limited modelling capabilities for complex topological structures, and cannot be applied to non-Euclidean spaces. On the contrary, graph neural networks (GNN) can well extract relational information in topological architecture and obtain high-level feature representation of images. In recent years, the modelling method based on graph structure has been applied in the 3D point cloud computing \cite{landrieu2018large, yue2019dynamic}, action recognition \cite{yan2018spatial, lin2014microsoft} and other fields. GNNs can solve image processing tasks that can be naturally constructed as graphs.
	
	\subsection{Graph Contrastive Learning}
	Graph contrastive learning methods aim to learn discriminative feature representations by enlarging the distance between positive and negative embedding samples. For example, Deep Graph Infomax (DGI) \cite{velickovic2019deep} enhances the feature representation ability of positively embedded nodes by maximizing the mutual information of global and local node representations. Graph Contrastive Adaptive Augmentation (GCA) \cite{zhu2021graph} strengthens the underlying semantic information in the graph by adding prior knowledge. Contrastive Multi-view Representation Learning (CMRL) \cite{hassani2020contrastive} obtains more discriminative feature representations by comparing feature representations of first-order neighbour nodes with node representations of graph diffusion.
	
	Nonetheless, the aforementioned graph-contrastive learning approach suffers from two problems. On the one hand, almost all methods need to construct multiple graph contrast views to generate positive and negative embedding samples, which is very computationally intensive. On the other hand, existing graph-contrastive learning methods cannot maintain a safe distance between intra-class and inter-class distances. To sum up, existing graph-contrastive learning methods suffer from high computational complexity and low discriminative power of feature representations.
	
	\section{Proposed Method}
	In this section, we propose a novel Masked Contrastive Graph Representation Learning (MCGRL) method for age estimation.
	
	\begin{figure*}
		\centering
		\includegraphics[width=1\linewidth]{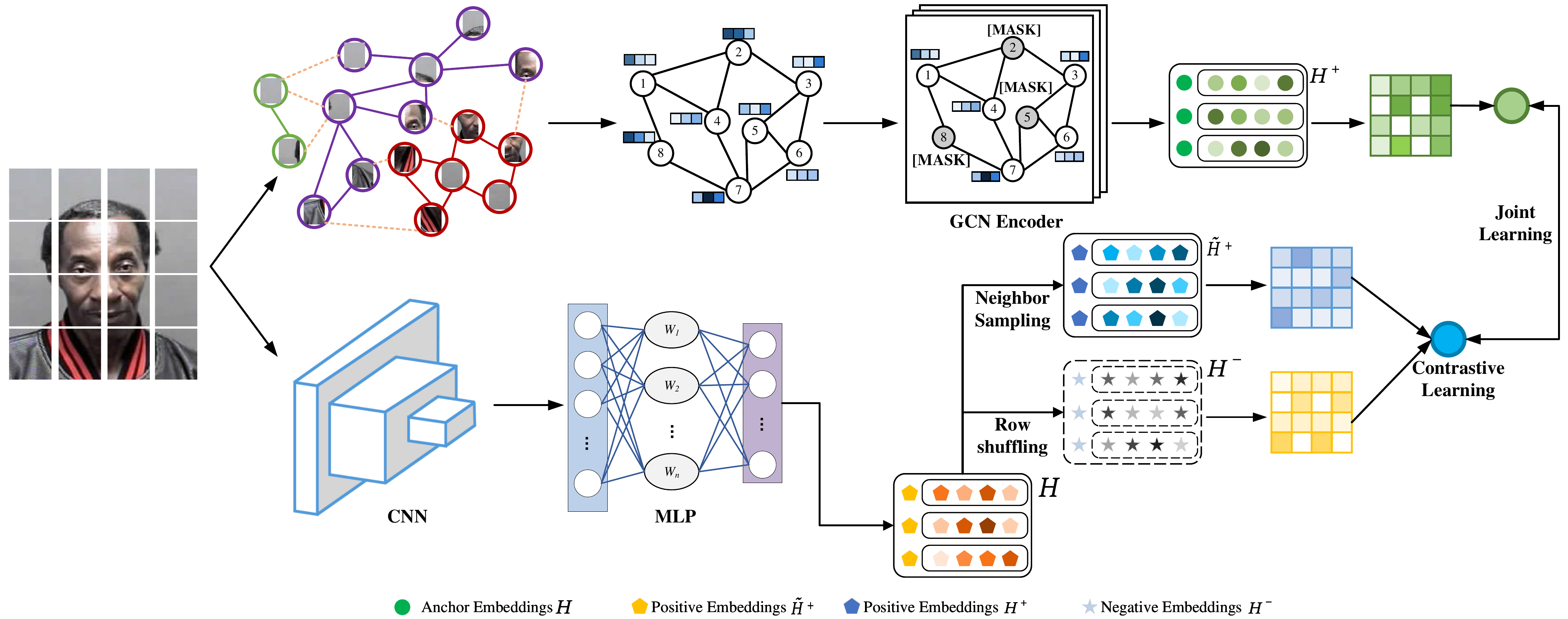}
		\caption{The flowchart of the proposed MCGRL method. Specifically, we first use masked GCN and CNN to extract structural information $H^+$ and semantic features $H$ in face images, respectively. $H$ is viewed as an anchor embedded sample. Then we use $H^+$ and the semantic features obtained $\tilde{H}^+$ by neighbour sampling as positive embedding samples, and the semantic information obtained $H^-$ by row-wise random permutation as negative samples. Finally, the positive samples are spatially close to the anchor embeddings and far away from the negative samples by performing multiple losses.}
		\label{fig:gnn}
	\end{figure*}

	\subsection{Graph Construction from Image} 
	First, we segment a face image of size $H \times W\times 3$ into $N$ patches. Each patch is transformed into a feature representation $\xi_i \in \mathbb{R}^d$, where $d$ represents the dimensionality of each feature representation and $i=\{1,2,\ldots,N\}$. Then, the feature representations of these patches are regarded as unordered nodes which are denoted as $\mathcal{V}=\{v_1,v_2,\ldots,v_N\}$. For each central node in the graph, we find its $K$ nearest neighbour nodes $\mathcal{N}(v_i)$ for edge building. Finally, we obtain a graph $\mathcal{G}=(\mathcal{V},\mathcal{E},\mathcal{R},\mathcal{W})$, where $\mathcal{E}$ is the set of all edges, the directed edge $r_{ij} (r_{ij}\in \mathcal{R})$ indicates that there is a connection relationship between node $v_i$ and node $v_j$. $\omega_{ij} (\omega_{ij}\in W, 0\leq \omega_{ij}\leq 1)$ represents the weight of edge $r_{ij}$, and $r\in \mathcal{R}$ is the relation type of the edge.

	\textbf{Mask generator:} We use mask operations on some nodes in the graph to improve the feature representation ability of GCN~\cite{kipfsemi}. Specifically, we first set an actual mask rate $p$. Then we generate an all-ones matrix of the same size as the input image and randomly set some elements to 0 according to the masking rate $p$. The specific formula of the mask generator is defined as follows:
	\begin{equation}
		\mathcal{V}^{[M]}=M^p \odot \mathcal{V},
	\end{equation}
	where $\mathcal{V}^{[M]}$ represents a masked graph, $M^p$ is a mask matrix, and $\odot$ represents a dot product operation. When $p$ is equal to 0, $M^p$ is an all-one matrix.

	\subsection{Anchor and Negative Embedding Generation}
	Previous work~\cite{cao2021bipartite, zeng2021contrastive} always conducts contrastive learning by constructing multiple graph contrastive views and uses the node representations aggregated by GCN as anchor embeddings. However, the computational complexity of GCN is very large, which will lead to a very long training time for the model. To speed up the computation, we use a CNN with MLP on the input face image to generate anchor embedding samples containing semantic features. The formula for anchor embedding generation is defined as follows:
	\begin{equation}
		\begin{gathered}
			\xi^{(l+1)}=\operatorname{Dropout}\left(\operatorname{LeakyReLU}\left(\operatorname{conv}\left(\xi^{(l+1)}\right) W^{(l)}\right)\right), \\
			H =\xi^{(l+1)} W^{(l+1)}.
		\end{gathered}
	\end{equation}
	where $\xi^0=\xi, W^{(l)}$ is the weight of layer $l$ in MLP.
	
	For the generation of negative embedding samples, we directly row-shuffle anchor embedding samples to obtain negative embedding samples. The formula is defined as follows:
	\begin{equation}
		H^{-}=\operatorname{Shuffle}\left(\left[\xi_1, \xi_2 \ldots, \xi_N\right]\right)
	\end{equation}
 
	\subsection{Positive Embedding Generation}	
	\subsubsection{Structural Information}
	GCN is used to obtain graph structure information of face images. To capture the key semantic information in nodes, we also introduce an attention mechanism to assign different weights to each edge. First, we utilize MLP to compute the correlation between node $i$ and node $j$. The formula is defined as follows:
	\begin{equation}
		\delta_{i j}^{(l+1)}=\operatorname{Dropout}\left(\operatorname{LeakyReLU}\left(W^{(l)}\left[\xi_i^{(l)} \oplus \xi_j^{(l)}\right]\right)\right),
	\end{equation}
	where $\oplus$ indicates the concatenation operation.
	
	Second, we use the softmax function to normalize the correlation coefficient and obtain the attention score of each edge. The formula is defined as follows:
	\begin{equation}
		\omega_{i j}^{(l+1)}=\operatorname{softmax}\left(\delta_{i j}^{(l+1)}\right)=\frac{\exp \left(\delta_{i j}^{(l+1)}\right)}{\sum_{\eta \in \mathcal{N}_i} \exp \left(\delta_{i j}^{(l+1)}\right)},
	\end{equation}
	where $\mathcal{N}_i$ represents the neighbor nodes of node $i$.
	
	Finally, node representations are updated by GCN with the LeakyReLU activation function. We take the updated structural information as positive embedding samples. The formula is defined as follows:
	\begin{equation}
		\begin{aligned}
			{H_{i}^{+}}^{(l+1)} &=\operatorname{LeakyReLU}\left(\sum_{r \in \mathcal{R}} \sum_{j \in \mathcal{N}_i^r} \frac{1}{\left|\mathcal{N}_i^r\right|}\left(\omega_{i j}^{(l)} W_{\theta_1}^{(l)} {H_{j}^{+}}^{(l)} \right.	\right.	\\ 
			&+ \left. \omega_{i i}^{(l)} W_{\theta_2}^{(l)} {H_{i}^{+}}^{(l)}\right)\Bigg),
		\end{aligned}
	\end{equation}
	
	\subsubsection{Neighbor Information}
	To obtain the neighbour information contained in the structure information, we sample the neighbour nodes of the central node and calculate their average value. In this way, we obtain the neighbour information of nodes as positive embedding samples:
	\begin{equation}
		\tilde{h}_j^{+}=\frac{1}{n} \sum_{j=1}^n\left\{h_j \mid v_j \in \mathcal{N}_i\right\} \text {, }
	\end{equation}
	where $n$ represents the number of samples.
	
	\subsection{Multiple Loss for Contrastive Learning}
	Different from previous work~\cite{zeng2021contrastive, cao2021bipartite} that constructs multiple graph contrasting views, we use triplet loss to compute the distance between anchor embeddings, positive embeddings and negative embeddings, and force the distance between positive embeddings to be closer and make the distance between negative embeddings and positive embeddings farther. In this way, the common characteristics among similar instances can be quickly learned. In addition, shrinking the intra-class variation and expanding the inter-class variation has also been proven to be an effective way to reduce the generalization error \cite{mo2022simple}. Specifically, the triplet loss is defined as follows:
	\begin{equation}
		\mathcal{L}_{t r i}=\frac{1}{m} \sum_{i=1}^m\left\{d\left(h, h^{+}\right)^2-d\left(h, h_i^{-}\right)^2+\alpha\right\}_{+},
	\end{equation}
	where $m$ represents the number of negative samples, $d(\cdot)$ is used to calculate the distance between samples and $\alpha$ is used to control the minimum distance between positive and negative samples, $\{\cdot\}=max\{\cdot,0\}$.
	
	Since Eq. (8) can only ensure that the distance between positive embedding and negative embedding can be enlarged without considering the inter-class difference, we thus construct two additional triplet losses to ensure that the inter-class differences can be enlarged. The formula is defined as follows:
	\begin{equation}
		\begin{aligned}
			& \mathcal{L}_N=\frac{1}{m} \sum_{i=1}^m\left\{d\left(h, h^{+}\right)^2-d\left(h, h_i^{-}\right)^2+\alpha\right\}_{+}, \\
			& \mathcal{L}_M=\frac{1}{m} \sum_{i=1}^m\left\{d\left(h, \tilde{h}^{+}\right)^2-d\left(h, h_i^{-}\right)^2+\alpha\right\}_{+}.
		\end{aligned}
	\end{equation}
	
	Since the structure information $h^+$ and neighbor information $\tilde{h}^+$ are different, when $d(h,h^+ )^2\geq d(h,\tilde{h}^+ )^2$, $\mathcal{L}_N$ is equal to 0 , $\mathcal{L}_M$ may not be equal to 0. In the above cases, $\mathcal{L}_N$ is not effective for the optimization of the model, while $\mathcal{L}_M$ is effective. When $d(h,h^+ )^2\leq d(h,\tilde{h}^+ )^2$, the conclusion also holds. Therefore, by combining the two cases in Eq. (9), the model can learn complementary semantic information between structural information $h^+$ and neighbour information $\tilde{h}^+$ and thus can enlarge the inter-class differences.

	The optimization goal of Eq. (8) is to make $d(h,h^+ )^2 - d(h,\tilde{h}^+ )^2$ closer to $\alpha$. In this case, the distance between the anchor embedding and the negative embedding is enlarged. However, there are cases where the distance between the anchor embedding and the positive embedding is also enlarged. For the above cases, we introduce upper bounds $\beta$ to ensure that the distances between anchor embeddings and negative embeddings, and also between anchor embeddings and positive embeddings, are within a controllable range. The formula is defined as follows:
	\begin{equation}
		\alpha+d\left\{h, h^{+}\right\}<d\left(h, h^{-}\right)<d\left(h, h^{+}\right)+\alpha+\beta,
	\end{equation}
	
	According to the analysis of Eq. (10), we can expand Eq. (8) as follows:
	\begin{equation}
		\mathcal{L}_V=-\frac{1}{m} \sum_{i=1}^m\left\{d\left(h, h^{+}\right)^2-d\left(h, h_i^{-}\right)^2+\alpha+\beta\right\}_{-},
	\end{equation}
	
	Finally, we combine the three loss functions of Eq. (9) and Eq. (11) as our multiple loss. The formula for multiple loss is defined as follows:
	\begin{equation}
		\mathcal{L}=\omega_1 \mathcal{L}_N+\omega_2 \mathcal{L}_M+\omega_3 \mathcal{L}_V,
	\end{equation}
	where $\omega_1, \omega_2, \omega_3$ are the hyper-parameters. In our experiments, they are empirically set as 1, 0.5, and 0.5, respectively.
	
	
	
	\subsection{Implementation Details}
	
	%
	We implement the model MCGRL proposed in this paper on a server with a Linux operating system and a graphics card of A100. The experimental environment of this paper is Python 3.7.0, and Pytorch 1.9.1. For the hyperparameter settings of the model, we specified a maximum number of iterations for the model of 50, a batch size of 196, a dropout of 0.5, a weight decay coefficient of 0.0001 and a learning rate of 1e-4. We set the minimum distance $\alpha$ and upper bound $\beta$ to 0.8 and 0.2, respectively.
	
	\section{Experiments}
	\subsection{Dataset and Evaluation Metrics}
	The MORPH-II\footnote{http://www.faceaginggroup.com/morph/}, FG-Net\footnote{http://yanweifu.github.io/FG\_NET\_data/FGNET.zip}, and CACD\footnote{http://bcsiriuschen.github.io/CARC/} datasets are widely used for age estimation. Therefore, this paper selects these three benchmark datasets to verify the effectiveness of our MCGRL method. The MORPH-II dataset contains 55,134 male and female face images, and each person has multiple images. The FGNET dataset consists of 1,002 face images ranging in age from 0 to 69, and each person has an average of 12 face images. Face images in the FGNET dataset are captured under different lighting and poses, so age estimation on this dataset is a challenging task. The CACD dataset consists of over 150000 images of 2000 celebrities collected from the internet. The dataset consists of a training set, a validation set, and a testing set, with 1800 people used for training, 80 people used for validation, and 120 people used for testing. To verify the generalization performance of the model, the FACES \footnote{http://faces.mpib-berlin.mpg.de}, SC-FACE \footnote{https://www.scface.org/}, and BAG datasets \footnote{http://multipie.org} are selected to verify the robustness of our MCGRL in cross-data evaluation. Two evaluation metrics, i.e., Mean Absolute Error (MAE) and Cumulative Score (CS), are chosen to evaluate the performance of each method. CS is the cumulative error of the model prediction results. CS represents the proportion of model prediction errors on the test set that are less than $L$ years.
	
	\begin{table}[htbp]
		\centering
		\caption{Experimental results of all the competing methods on the MORPH dataset. We count cumulative scores for age errors within a 5-year range. The best results in each column are in bold.}
		\renewcommand\arraystretch{1.2}
		\setlength{\tabcolsep}{20.5pt}{
			\begin{tabular}{lcc}
				\toprule
				\textbf{Method} & MAE  &  CS (\%)	\\ \midrule
				Human workers \cite{han2014demographic}   & 6.30 &  65.3	\\
				Ranking-CNN \cite{chen2017deep}    & 2.96 &  83.7    	\\
				DLDL \cite{gao2017deep}            & 2.42 &  87.4	\\
				CSOHR \cite{chang2015learning}          & 3.82 &  79.6	\\
				DEX  \cite{rothe2018deep}           & 2.68 &  84.9	\\
				Hu et al. \cite{hu2016facial}      & 2.78 &  82.1	\\
				Niu et al. \cite{niu2016ordinal}     & 3.27 &  74.3	\\
				CNN+ELM \cite{duan2017ensemble}        & 4.03 &  70.1	\\
				RAGN  \cite{duan2017ensemble}          & 2.61 &  86.6	\\ \hline
				MCGRL(ours)  &  \textbf{2.39}     &  \textbf{89.9}  \\      \bottomrule
		\end{tabular}}
	\end{table}
	
	\begin{table}[!t]
		\centering
		\caption{Experimental results of all the competing methods on the FG-NET dataset. We count cumulative scores for age errors within a 5-year range. The best results in each column are in bold.}
		\renewcommand\arraystretch{1.2}
		\setlength{\tabcolsep}{21pt}{
			\begin{tabular}{lcc}
				\toprule
				\textbf{Method} & MAE  & CS (\%) \\ \midrule
				Human workers \cite{han2014demographic}  & 4.70 & 69.5   \\
				Ranking-CNN \cite{chen2017deep}    & 5.79 & 66.5   \\
				DIF  \cite{han2014demographic}           & 4.80 & 74.3   \\
				AGES  \cite{geng2007automatic}          & 6.77 & 64.1   \\
				IIS-LDL  \cite{10.5555/2898607.2898680}       & 5.77 & 66.8    \\
				CPNN   \cite{geng2013facial}         & 4.76 & 68.1    \\
				MTWGP \cite{zhang2010multi}          & 4.83 & 72.3   \\
				CA-SVR  \cite{chen2013cumulative}        & 4.67 & 74.5   \\
				OHRank  \cite{chang2011ordinal}        & 4.48 & 74.4   \\
				DLA    \cite{wang2015deeply}         & 4.26 & 74.3    \\
				CAM    \cite{luu2011contourlet}         & 4.12 & 73.5   \\
				Rothe \cite{rothe2016some}          & 5.01 & 67.5    \\
				DEX   \cite{rothe2018deep}          & 4.63 & 72.4    \\
				DRF \cite{shen2019deep}            & 3.47 & 77.3   \\ 
				VDAL	\cite{liu2020similarity}		&  2.98	& 81.9   \\    \hline
				MCGRL (ours)  &   \textbf{2.86}    & \textbf{88.0}   \\  \bottomrule
		\end{tabular}}
	\end{table}
	
	\begin{table}[htbp]
		\centering
		\caption{Experimental results of all the competing methods on the CACD dataset. We count cumulative scores for age errors within a 5-year range. The best results in each column are in bold.}
		\renewcommand\arraystretch{1.2}
		\setlength{\tabcolsep}{21.8pt}{
			\begin{tabular}{lcc}
				\toprule
				Methods & MAE  & CS (\%)  \\ \midrule
				DEX  \cite{rothe2018deep}                       & 6.52  & 68.2  \\
				DLDLF \cite{shen2019deep}                      & 6.16  & 71.1 \\
				DRF  \cite{shen2019deep}                       & 5.63  & 72.8 \\ \hline
				MCGRL (ours)  &          \textbf{4.03}   &  \textbf{80.1}     \\   \bottomrule
		\end{tabular}}
	\end{table}
	
	\begin{figure*}
		\centering
		\includegraphics[width=1\linewidth]{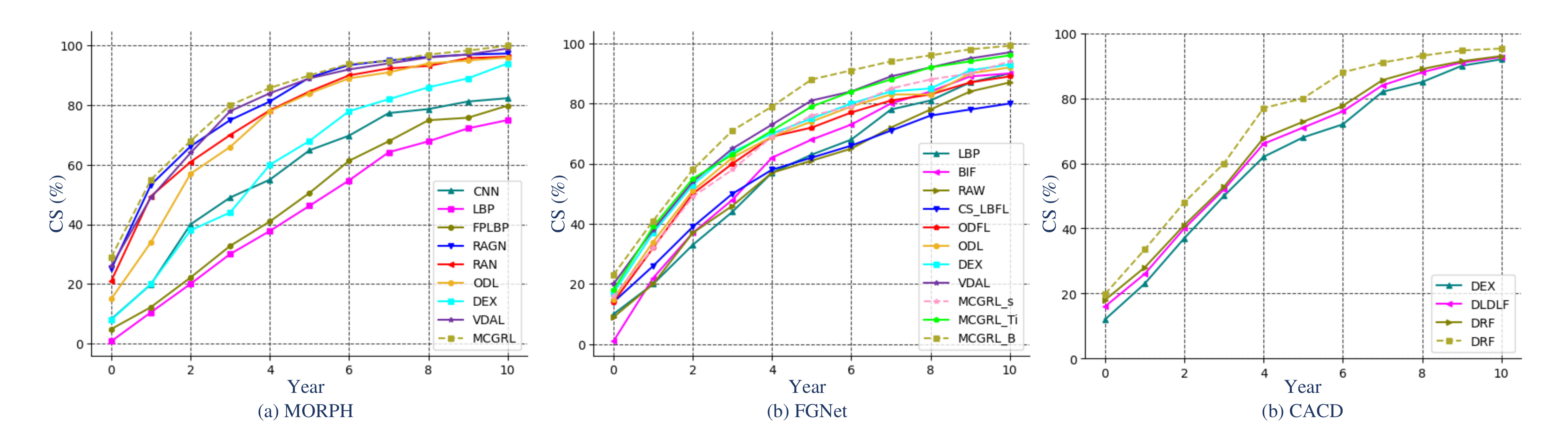}
		\caption{We compare the cumulative scores of different methods on the MORPH, FGNET and CACD datasets, respectively. MCGRL achieves the best results in cumulative scores across all age error ranges.}
		\label{fig:score}
	\end{figure*}
	
	\begin{figure*}
		\centering
		\includegraphics[width=1\linewidth]{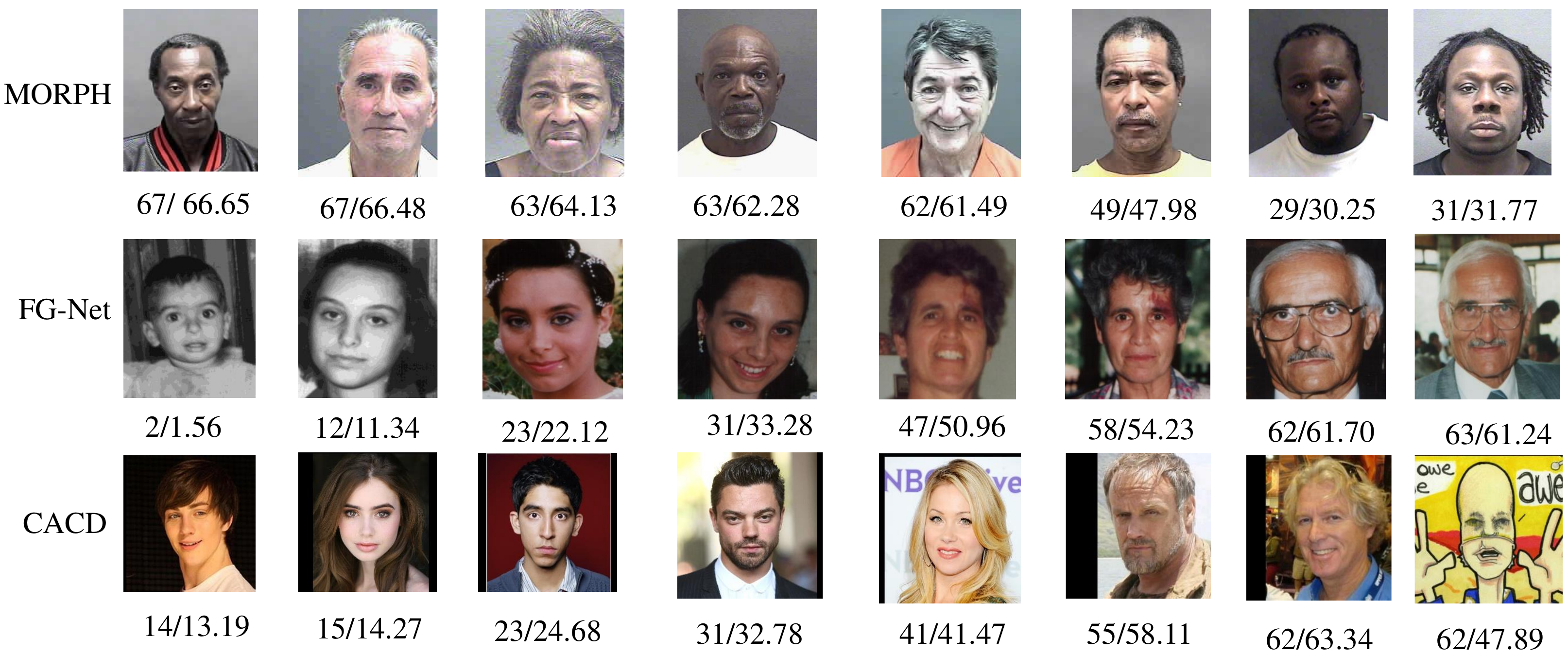}
		\caption{Prediction results (GT/Pred) of our MCGRL method on some sample images of the three benchmark datasets.}
		\label{fig:predict}
	\end{figure*}
	
	\subsection{Performance Verification Experiment}
	To verify the effectiveness of our proposed MCGRL method, we test our method on three benchmark datasets of face images. The experimental results are shown in Table 1, Table 2 and Table 3. All three variants of our MCGRL are on par or ahead of the state-of-the-art methods, especially on the CACD dataset. We conjecture that the improved performance is due to the flexible modelling of irregular face images by the graph structure. However, existing comparison methods can only use regular grid structures or sequence structures to model face images, which contain redundant information. In addition, we also compare the results of cumulative scores under different years of error as shown in Figure~\ref{fig:score}. MCGRL maintains the best cumulative score at each stage.

	\subsection{Visualization of Prediction Results}
	We randomly select 8 images from the test set in the three datasets MORPH, FGNET and CACD for age prediction respectively. As shown in Figure~\ref{fig:predict}, the prediction error of MCGRL on most face images is less than 1 year old, while the error between the prediction results and the real results on a small number of complex images is relatively large. Overall, the prediction results of MCGRL on the test sets are reliable.
	
	\begin{table*}[htbp]
		\centering
		\caption{Cross-dataset evaluation method to verify the generalization performance of the model (Training data: BAG-Full). We count cumulative scores for age errors within a 5-year range. The best results in each column are in bold.}
		\renewcommand\arraystretch{1.2}
		\setlength{\tabcolsep}{7.5pt}{
			\begin{tabular}{lcccccccccccc}
				\toprule
				\multirow{2}{*}{\textbf{Method}} & \multicolumn{2}{c}{FG-NET} & \multicolumn{2}{c}{MORPH} & \multicolumn{2}{c}{FACES} & \multicolumn{2}{c}{SC-FACE-ROT} & \multicolumn{2}{c}{SC-FACE-SUR} & \multicolumn{2}{c}{Average} \\ \cline{2-13} 
				& MAE         & CS(\%)       & MAE        & CS(\%)       & MAE        & CS(\%)       & MAE           & CS(\%)          & MAE           & CS(\%)          & MAE         & CS(\%)        \\ \midrule
				Human Workers                    & 4.70        & 69.50        & 6.30       & 51.0         & NA         & NA           & NA            & NA              & NA            & NA              & 5.50        & 60.25         \\
				Hard-Ranking                     & 3.74        & 78.50        & 5.48       & 60.25        & 5.22       & 61.20        & 5.15          & 70.25           & 5.59          & 62.01           & 5.03        & 66.44         \\
				DEX                              & 3.20        & 82.14        & 5.50       & 60.34        & 5.33       & 61.60        & 6.07          & 53.59           & 5.44          & 66.76           & 5.10        & 64.88         \\
				AGEn                             & 3.15        & 82.98        & 5.40       & 60.95        & 5.13       & 63.12        & 5.84          & 55.98           & 5.32          & 67.02           & 4.96        & 66.01         \\
				Soft-Ranking                     & 3.12        & 83.80        & 5.28       & 62.55        & 4.83       & 65.74        & 5.29          & 63.92           & 5.41          & 64.90           & 4.78        & 68.18         \\
				DLDL                             & 3.08        & 83.83        & 5.27       & 62.43        & 4.72       & 66.76        & 5.25          & 63.93           & 5.46          & 65.71           & 4.75        & 68.53         \\
				CE-MV                            & 3.07        & 83.23        & 5.22       & 61.31        & 4.62       & 69.88        & 4.89          & 64.35           & 5.13          & 69.98           & 4.58        & 69.75         \\
				DLDL-v2                          & 3.06        & 82.83        & 4.95       & 64.95        & 4.39       & 71.00        & 5.24          & 64.70           & 4.90          & 71.27           & 4.50        & 70.95         \\
				DC                               & 2.93        & 84.43        & 4.63       & 66.03        & 4.47       & 69.88        & 4.72          & 71.19           & 4.78          & 71.75           & 4.30        & 72.65       \\  
				MCGRL (ours)  &  \textbf{2.67}     & \textbf{85.69}   & \textbf{4.27}  &  \textbf{69.12}   &   \textbf{3.80}  &  \textbf{76.23}  &  \textbf{4.01}  &   \textbf{76.20}  &  \textbf{4.00}   & \textbf{76.66}  & \textbf{3.75}  & \textbf{76.78}   \\ 
				\bottomrule
		\end{tabular}}
	\end{table*}
	
	
	\begin{table*}[htbp]
		\centering
		\caption{Cross-dataset evaluation method to verify the generalization performance of the model (Training data: BAG-Subset). We count cumulative scores for age errors within a 5-year range. The best results in each column are in bold.}
		\renewcommand\arraystretch{1.2}
		\setlength{\tabcolsep}{7.5pt}{
			\begin{tabular}{lcccccccccccc}
				\toprule
				\multirow{2}{*}{\textbf{Method}} & \multicolumn{2}{c}{FG-NET} & \multicolumn{2}{c}{MORPH} & \multicolumn{2}{c}{FACES} & \multicolumn{2}{c}{SC-FACE-ROT} & \multicolumn{2}{c}{SC-FACE-SUR} & \multicolumn{2}{c}{Average} \\ \cline{2-13} 
				& MAE         & CS(\%)       & MAE        & CS(\%)       & MAE        & CS(\%)       & MAE           & CS(\%)          & MAE           & CS(\%)          & MAE         & CS(\%)        \\ \midrule
				DEX                              & 3.57        & 78.94        & 6.54       & 53.38        & 6.59       & 50.83        & 6.45          & 49.32           & 6.19          & 65.05           & 5.86        & 59.50         \\
				AGEn                             & 3.53        & 79.78        & 6.40       & 53.97        & 6.34       & 52.40        & 6.25          & 51.65           & 6.12          & 65.21           & 5.72        & 60.60         \\
				DLDL                             & 3.24        & 81.54        & 6.01       & 57.36        & 6.11       & 55.60        & 5.90          & 54.79           & 6.52          & 60.64           & 5.55        & 61.98         \\
				CE-MV                            & 3.34        & 80.44        & 6.22       & 55.60        & 6.25       & 54.63        & 6.06          & 54.19           & 6.23          & 64.38           & 5.62        & 61.84         \\
				DLDL-v2                          & 3.35        & 81.44        & 5.80       & 57.30        & 5.92       & 56.68        & 5.82          & 56.84           & 6.52          & 61.61           & 5.48        & 62.77         \\
				DC                               & 3.21        & 81.59        & 5.63       & 59.13        & 5.90       & 57.55        & 5.32          & 62.14           & 5.37          & 67.96           & 5.08        & 65.67     \\    
				MCGRL (ours)  &   \textbf{3.01}    &  \textbf{83.28}  &  \textbf{5.44} &   \textbf{61.34}  &   \textbf{5.70}  &  \textbf{59.23}  &  \textbf{5.13}  &  \textbf{63.21}   &   \textbf{4.87}  & \textbf{69.27}  & \textbf{4.83}  &  \textbf{67.28}  \\  
				\bottomrule
		\end{tabular}}
	\end{table*}
	
	
	\begin{table*}[htbp]
		\centering
		\caption{Cross-dataset evaluation method to verify the generalization performance of the model (Training data: MORPH). The best results in each column are in bold.}
		\renewcommand\arraystretch{1.2}
		\setlength{\tabcolsep}{11pt}{
			\begin{tabular}{lcccccccccc}
				\toprule
				\multirow{2}{*}{\textbf{Method}} & \multicolumn{2}{c}{FG-NET} & \multicolumn{2}{c}{FACES} & \multicolumn{2}{c}{SC-FACE-ROT} & \multicolumn{2}{c}{SC-FACE-SUR} & \multicolumn{2}{c}{Average} \\ \cline{2-11} 
				& MAE         & CS(\%)       & MAE        & CS(\%)       & MAE           & CS(\%)          & MAE           & CS(\%)          & MAE         & CS(\%)        \\ \hline
				DEX                              & 5.73        & 58.31        & 9.28       & 36.53        & 4.68          & 67.95           & 9.14          & 23.68           & 7.20        & 46.61         \\
				DLDL                             & 5.45        & 62.76        & 8.43       & 40.90        & 4.22          & 71.54           & 9.70          & 23.96           & 6.95        & 49.79         \\
				DC                               & 5.29        & 63.70        & 8.73       & 40.45        & 4.05          & 76.24           & 8.92          & 29.40           & 6.74        & 52.44   \\
				MCGRL (ours)  &   \textbf{4.34}    &  \textbf{73.59}  & \textbf{6.50}  & \textbf{57.21 }   &  \textbf{3.61}   &  \textbf{79.88}  & \textbf{ 6.53}  &  \textbf{48.37}   &  \textbf{5.25}   &   \textbf{64.76}    \\  \bottomrule
		\end{tabular}}
	\end{table*}
	
	\subsection{Comparison of CNN and ViT methods}
\begin{figure}
		\centering
		\includegraphics[width=1\linewidth]{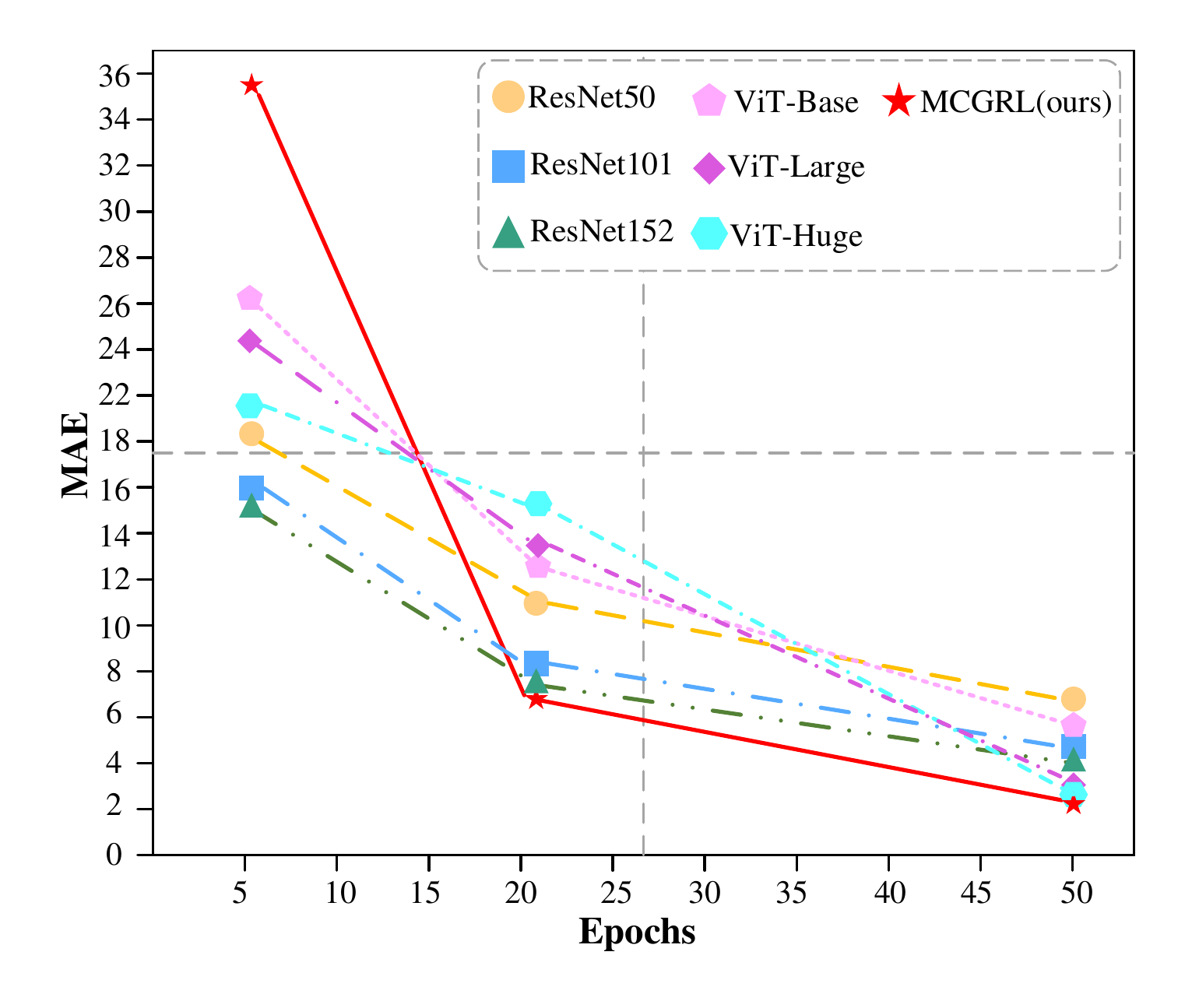}
		\caption{We test the trained MCGRL model on the validation set to obtain the predicted age error. The MAE value of MCGRL at the beginning of training is much higher than that of CNN and ViT. As the epoch number of training increases, the MAE value of MCGRL is gradually lower than CNN and VIT.}
		\label{fig:training}
	\end{figure}

	To illustrate the superiority of our method MCGRL more intuitively, we compared it with the traditional CNN method based on the grid structure and the Vision Transformer method based on the sequence structure (i.e., ResNet18, ResNet18, ResNet18, ViT-Base, ViT- Large, and ViT-Large). As shown in Figure \ref{fig:training}, the MAE value of the MCGRL method in the initial training process is much higher than the ResNet and ViT methods, but as the number of training increases, the convergence speed of MCGRL far exceeds other methods. The improvement of the convergence speed may be attributed to the design of the mask contrastive learning mechanism, which can quickly learn similar features of the same kind of data from a large amount of data and encode them into advanced representations. While other methods outperform MCGRL in predicting the initial training process, it may be attributed to the powerful feature representation ability of the pre-trained model. Nevertheless, MCGRL outperforms CNN and ViT finally, we think this is because MCGRL is more flexible in modelling face images and can remove redundant information in images. Therefore, applying graph neural networks to the field of age estimation from face images has a potential application.
	
	\subsection{Generalization Performance Verification}
	Since the evaluation of model performance within the dataset cannot reflect the generalization performance of the model, we use a cross-dataset evaluation method to verify the generalization performance of our model. The experimental results are shown in Tables 4, 5, and 6. Our method performs significantly better than other algorithms on different data sets. The improvement in generalization performance may be attributed to the fact that the graph structure can flexibly model any complex image, while other methods contain redundant information when extracting features from irregular objects.
	
	\begin{table}[htbp]
		\centering
		\caption{Different GNNs are used to verify their impact on the experimental results. Experimental results are compared on the MORPH, FG-NET, and CACD datasets. MAE is chosen as our evaluation metric. The best results in each column are in bold.}
		\renewcommand\arraystretch{1.2}
		\setlength{\tabcolsep}{7pt}{
			\begin{tabular}{lccc}
				\toprule
				GraphConv Type & MORPH & FG-NET & CACD \\ \midrule
				EdgeConv                                         &   2.43    &    2.91    & 4.19     \\
				GIN                                               &   2.66    &   3.17     &   4.43   \\
				GraphSAGE                                         &   2.57    &    3.08    &  4.32    \\
				Max-Relative GraphConv                           &   \textbf{2.39}    &    \textbf{2.86}    &   \textbf{4.03}   \\ \bottomrule
		\end{tabular}}
	\end{table}

	\subsection{Ablation Study}
	We perform an ablation study to verify the effectiveness of each module of our proposed method on MORPH, FGNET and CACD datasets and use MCGRL-B as our network architecture.
	
		\begin{table}[!t]
		\centering
		\caption{We conduct ablation experiments on triplet loss ($\mathcal{L}_N$ and $\mathcal{L}_M$) and upper bound loss ($\mathcal{L}_V$) to verify their effectiveness. The experimental results are compared on the MORPH, FG-NET, and CACD datasets. The best results in each column are in bold.}
		\renewcommand\arraystretch{1.2}
		\setlength{\tabcolsep}{9pt}{
			\begin{tabular}{cccccc}
				\toprule
				$\mathcal{L}_N$ & $\mathcal{L}_M$ & $\mathcal{L}_V$ & MORPH & FG-NET & CACD \\ \midrule
				-          & - & - & 5.34     & 5.87      & 6.97    \\
				$\surd$          & - & - & 4.64     & 4.79      & 5.66    \\
				-          & $\surd$ & - & 4.12     & 4.27      & 5.59    \\
				$\surd$         & $\surd$ & - & 4.03     & 4.05      & 5.02    \\
				$\surd$          & - & $\surd$ & 3.74     & 3.47      & 4.68    \\
				-          & $\surd$ & $\surd$ & 2.62     & 2.96     & 4.43   \\
				$\surd$          & $\surd$ & $\surd$ & \textbf{2.39}     & \textbf{2.86}     & \textbf{4.03}   \\ \bottomrule
		\end{tabular}}
	\end{table}
	
	\subsubsection{Type of Graph Convolution}
	We test the feature representation capabilities of different variants of graph convolutional neural networks, i.e., Max-Relative GraphConv, EdgeConv, GraphSAGE, and GIN. As shown in 7, Max-Relative GraphConv achieves the best MAE on all three datasets. Therefore, we use Max-Relative GraphConv by default in our experiments.
	
	\subsubsection{Necessity of Triple Loss}
	We perform ablation experiments to analyze the effect of triplet loss ($\mathcal{L}_N$ and $\mathcal{L}_M$) and upper bound loss ($\mathcal{L}_V$) on model performance. As shown in Table 8, the model achieves the best experimental results when all losses are included. When no modules are used (i.e. only GCNs are used for age prediction), the prediction results of the model are worst. If only $\mathcal{L}_V$ is used, the contrast loss cannot be formed. If only $\mathcal{L}_N$ or $\mathcal{L}_M$ is used, the prediction results of the model are relatively poor. On the contrary, any combination of two losses can improve the prediction accuracy of the model.

	\subsubsection{Hyper-parameter Analysis}
\begin{figure}
		\centering
		\includegraphics[width=1\linewidth]{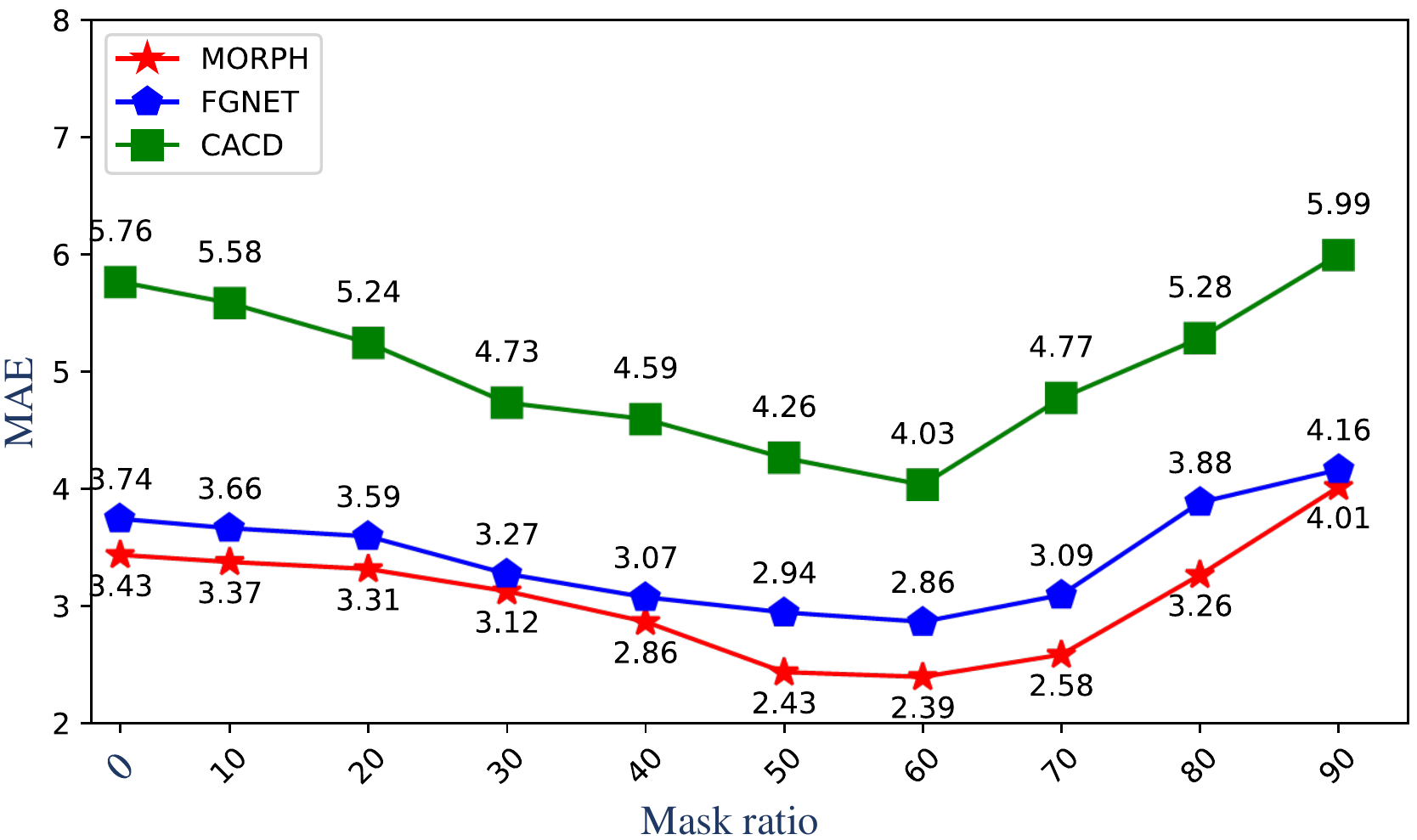}
		\caption{Performance of different mask ratios on three datasets. When the mask rate $p$ is less than 0.6, as $p$ increases, the MAE value gradually decreases and reaches the optimal value when $p=0.6$. When $p>0.6$, the semantic information of the graph is seriously lost and the MAE value begins to decrease.}
		\label{fig:agemask}
	\end{figure}

	We investigate the effect of hyperparameters on MCGRL, i.e., the masking rate $p$. As shown in Figure \ref{fig:agemask}, we set the mask rate $p$ from 0.1 to 0.9, and MCGRL-B can achieve the best experimental results when the mask rate is 0.6. If the masking rate $p$ is too large, the graph structure loses a large amount of semantic information, so that the features of the nodes cannot be restored.
	
	\section{Conclusion}
	In this paper, a novel Masked Contrastive Graph Representation Learning for Age Estimation architecture, named MCGRL, is proposed. Unlike previous architectures using CNN for feature extraction, MCGRL adopts a more flexible GNN to capture irregular and complex objects in images. In order to improve the fusion representation ability of node features and structures in graphs, we design a self-supervised masked graph autoencoder (SMGAE) to perform mask reconstruction on nodes. The feature vectors encoded and decoded by SMGAE have stronger semantic representation ability. Furthermore, to widen the difference between different classes and narrow the gap between the same classes, we introduce a contrastive learning mechanism to improve the generalization performance of the model. On four benchmark datasets for age estimation, MCGRL outperforms existing comparison algorithms.
	
	
	\bibliographystyle{ACM-Reference-Format}
	\bibliography{refs}


\begin{thebibliography}{53}


\ifx \showCODEN    \undefined \def \showCODEN     #1{\unskip}     \fi
\ifx \showDOI      \undefined \def \showDOI       #1{#1}\fi
\ifx \showISBNx    \undefined \def \showISBNx     #1{\unskip}     \fi
\ifx \showISBNxiii \undefined \def \showISBNxiii  #1{\unskip}     \fi
\ifx \showISSN     \undefined \def \showISSN      #1{\unskip}     \fi
\ifx \showLCCN     \undefined \def \showLCCN      #1{\unskip}     \fi
\ifx \shownote     \undefined \def \shownote      #1{#1}          \fi
\ifx \showarticletitle \undefined \def \showarticletitle #1{#1}   \fi
\ifx \showURL      \undefined \def \showURL       {\relax}        \fi
\providecommand\bibfield[2]{#2}
\providecommand\bibinfo[2]{#2}
\providecommand\natexlab[1]{#1}
\providecommand\showeprint[2][]{arXiv:#2}

\bibitem[\protect\citeauthoryear{Allen-Zhu and Li}{Allen-Zhu and Li}{2019}]%
        {allen2019can}
\bibfield{author}{\bibinfo{person}{Zeyuan Allen-Zhu} {and}
  \bibinfo{person}{Yuanzhi Li}.} \bibinfo{year}{2019}\natexlab{}.
\newblock \showarticletitle{What can resnet learn efficiently, going beyond
  kernels?}
\newblock \bibinfo{journal}{\emph{Advances in Neural Information Processing
  Systems}}  \bibinfo{volume}{32} (\bibinfo{year}{2019}).
\newblock


\bibitem[\protect\citeauthoryear{Arnab, Dehghani, Heigold, Sun,
  Lu{\v{c}}i{\'c}, and Schmid}{Arnab et~al\mbox{.}}{2021}]%
        {arnab2021vivit}
\bibfield{author}{\bibinfo{person}{Anurag Arnab}, \bibinfo{person}{Mostafa
  Dehghani}, \bibinfo{person}{Georg Heigold}, \bibinfo{person}{Chen Sun},
  \bibinfo{person}{Mario Lu{\v{c}}i{\'c}}, {and} \bibinfo{person}{Cordelia
  Schmid}.} \bibinfo{year}{2021}\natexlab{}.
\newblock \showarticletitle{Vivit: A video vision transformer}. In
  \bibinfo{booktitle}{\emph{Proceedings of the IEEE/CVF International
  Conference on Computer Vision}}. \bibinfo{pages}{6836--6846}.
\newblock


\bibitem[\protect\citeauthoryear{Cao, Lei, Zhang, Feng, and Li}{Cao
  et~al\mbox{.}}{2012}]%
        {cao2012human}
\bibfield{author}{\bibinfo{person}{Dong Cao}, \bibinfo{person}{Zhen Lei},
  \bibinfo{person}{Zhiwei Zhang}, \bibinfo{person}{Jun Feng}, {and}
  \bibinfo{person}{Stan~Z Li}.} \bibinfo{year}{2012}\natexlab{}.
\newblock \showarticletitle{Human age estimation using ranking svm}. In
  \bibinfo{booktitle}{\emph{Biometric Recognition: 7th Chinese Conference, CCBR
  2012, Guangzhou, China, December 4-5, 2012. Proceedings 7}}. Springer,
  \bibinfo{pages}{324--331}.
\newblock


\bibitem[\protect\citeauthoryear{Cao, Lin, Guo, Liu, Liu, and Wang}{Cao
  et~al\mbox{.}}{2021}]%
        {cao2021bipartite}
\bibfield{author}{\bibinfo{person}{Jiangxia Cao}, \bibinfo{person}{Xixun Lin},
  \bibinfo{person}{Shu Guo}, \bibinfo{person}{Luchen Liu},
  \bibinfo{person}{Tingwen Liu}, {and} \bibinfo{person}{Bin Wang}.}
  \bibinfo{year}{2021}\natexlab{}.
\newblock \showarticletitle{Bipartite graph embedding via mutual information
  maximization}. In \bibinfo{booktitle}{\emph{Proceedings of the 14th ACM
  International Conference on Web Search and Data Mining}}.
  \bibinfo{pages}{635--643}.
\newblock


\bibitem[\protect\citeauthoryear{Cao, Mirjalili, and Raschka}{Cao
  et~al\mbox{.}}{2020}]%
        {cao2020rank}
\bibfield{author}{\bibinfo{person}{Wenzhi Cao}, \bibinfo{person}{Vahid
  Mirjalili}, {and} \bibinfo{person}{Sebastian Raschka}.}
  \bibinfo{year}{2020}\natexlab{}.
\newblock \showarticletitle{Rank consistent ordinal regression for neural
  networks with application to age estimation}.
\newblock \bibinfo{journal}{\emph{Pattern Recognition Letters}}
  \bibinfo{volume}{140} (\bibinfo{year}{2020}), \bibinfo{pages}{325--331}.
\newblock


\bibitem[\protect\citeauthoryear{Chang and Chen}{Chang and Chen}{2015}]%
        {chang2015learning}
\bibfield{author}{\bibinfo{person}{Kuang-Yu Chang} {and}
  \bibinfo{person}{Chu-Song Chen}.} \bibinfo{year}{2015}\natexlab{}.
\newblock \showarticletitle{A learning framework for age rank estimation based
  on face images with scattering transform}.
\newblock \bibinfo{journal}{\emph{IEEE Transactions on Image Processing}}
  \bibinfo{volume}{24}, \bibinfo{number}{3} (\bibinfo{year}{2015}),
  \bibinfo{pages}{785--798}.
\newblock


\bibitem[\protect\citeauthoryear{Chang, Chen, and Hung}{Chang
  et~al\mbox{.}}{2011}]%
        {chang2011ordinal}
\bibfield{author}{\bibinfo{person}{Kuang-Yu Chang}, \bibinfo{person}{Chu-Song
  Chen}, {and} \bibinfo{person}{Yi-Ping Hung}.}
  \bibinfo{year}{2011}\natexlab{}.
\newblock \showarticletitle{Ordinal hyperplanes ranker with cost sensitivities
  for age estimation}. In \bibinfo{booktitle}{\emph{CVPR 2011}}. IEEE,
  \bibinfo{pages}{585--592}.
\newblock


\bibitem[\protect\citeauthoryear{Chen, Gong, Xiang, and Change~Loy}{Chen
  et~al\mbox{.}}{2013}]%
        {chen2013cumulative}
\bibfield{author}{\bibinfo{person}{Ke Chen}, \bibinfo{person}{Shaogang Gong},
  \bibinfo{person}{Tao Xiang}, {and} \bibinfo{person}{Chen Change~Loy}.}
  \bibinfo{year}{2013}\natexlab{}.
\newblock \showarticletitle{Cumulative attribute space for age and crowd
  density estimation}. In \bibinfo{booktitle}{\emph{Proceedings of the IEEE
  conference on computer vision and pattern recognition}}.
  \bibinfo{pages}{2467--2474}.
\newblock


\bibitem[\protect\citeauthoryear{Chen, Zhang, and Dong}{Chen
  et~al\mbox{.}}{2017}]%
        {chen2017deep}
\bibfield{author}{\bibinfo{person}{Shixing Chen}, \bibinfo{person}{Caojin
  Zhang}, {and} \bibinfo{person}{Ming Dong}.} \bibinfo{year}{2017}\natexlab{}.
\newblock \showarticletitle{Deep age estimation: From classification to
  ranking}.
\newblock \bibinfo{journal}{\emph{IEEE Transactions on Multimedia}}
  \bibinfo{volume}{20}, \bibinfo{number}{8} (\bibinfo{year}{2017}),
  \bibinfo{pages}{2209--2222}.
\newblock


\bibitem[\protect\citeauthoryear{Duan, Li, and Li}{Duan et~al\mbox{.}}{2017}]%
        {duan2017ensemble}
\bibfield{author}{\bibinfo{person}{Mingxing Duan}, \bibinfo{person}{Kenli Li},
  {and} \bibinfo{person}{Keqin Li}.} \bibinfo{year}{2017}\natexlab{}.
\newblock \showarticletitle{An ensemble CNN2ELM for age estimation}.
\newblock \bibinfo{journal}{\emph{IEEE Transactions on Information Forensics
  and Security}} \bibinfo{volume}{13}, \bibinfo{number}{3}
  (\bibinfo{year}{2017}), \bibinfo{pages}{758--772}.
\newblock


\bibitem[\protect\citeauthoryear{Gao, Xing, Xie, Wu, and Geng}{Gao
  et~al\mbox{.}}{2017}]%
        {gao2017deep}
\bibfield{author}{\bibinfo{person}{Bin-Bin Gao}, \bibinfo{person}{Chao Xing},
  \bibinfo{person}{Chen-Wei Xie}, \bibinfo{person}{Jianxin Wu}, {and}
  \bibinfo{person}{Xin Geng}.} \bibinfo{year}{2017}\natexlab{}.
\newblock \showarticletitle{Deep label distribution learning with label
  ambiguity}.
\newblock \bibinfo{journal}{\emph{IEEE Transactions on Image Processing}}
  \bibinfo{volume}{26}, \bibinfo{number}{6} (\bibinfo{year}{2017}),
  \bibinfo{pages}{2825--2838}.
\newblock


\bibitem[\protect\citeauthoryear{Geng, Smith-Miles, and Zhou}{Geng
  et~al\mbox{.}}{2010}]%
        {10.5555/2898607.2898680}
\bibfield{author}{\bibinfo{person}{Xin Geng}, \bibinfo{person}{Kate
  Smith-Miles}, {and} \bibinfo{person}{Zhi-Hua Zhou}.}
  \bibinfo{year}{2010}\natexlab{}.
\newblock \showarticletitle{Facial Age Estimation by Learning from Label
  Distributions}. In \bibinfo{booktitle}{\emph{Proceedings of the Twenty-Fourth
  AAAI Conference on Artificial Intelligence}}. \bibinfo{publisher}{AAAI
  Press}, \bibinfo{pages}{451–456}.
\newblock


\bibitem[\protect\citeauthoryear{Geng, Yin, and Zhou}{Geng
  et~al\mbox{.}}{2013}]%
        {geng2013facial}
\bibfield{author}{\bibinfo{person}{Xin Geng}, \bibinfo{person}{Chao Yin}, {and}
  \bibinfo{person}{Zhi-Hua Zhou}.} \bibinfo{year}{2013}\natexlab{}.
\newblock \showarticletitle{Facial age estimation by learning from label
  distributions}.
\newblock \bibinfo{journal}{\emph{IEEE transactions on pattern analysis and
  machine intelligence}} \bibinfo{volume}{35}, \bibinfo{number}{10}
  (\bibinfo{year}{2013}), \bibinfo{pages}{2401--2412}.
\newblock


\bibitem[\protect\citeauthoryear{Geng, Zhou, and Smith-Miles}{Geng
  et~al\mbox{.}}{2007}]%
        {geng2007automatic}
\bibfield{author}{\bibinfo{person}{Xin Geng}, \bibinfo{person}{Zhi-Hua Zhou},
  {and} \bibinfo{person}{Kate Smith-Miles}.} \bibinfo{year}{2007}\natexlab{}.
\newblock \showarticletitle{Automatic age estimation based on facial aging
  patterns}.
\newblock \bibinfo{journal}{\emph{IEEE Transactions on pattern analysis and
  machine intelligence}} \bibinfo{volume}{29}, \bibinfo{number}{12}
  (\bibinfo{year}{2007}), \bibinfo{pages}{2234--2240}.
\newblock


\bibitem[\protect\citeauthoryear{Han, Otto, Liu, and Jain}{Han
  et~al\mbox{.}}{2014}]%
        {han2014demographic}
\bibfield{author}{\bibinfo{person}{Hu Han}, \bibinfo{person}{Charles Otto},
  \bibinfo{person}{Xiaoming Liu}, {and} \bibinfo{person}{Anil~K Jain}.}
  \bibinfo{year}{2014}\natexlab{}.
\newblock \showarticletitle{Demographic estimation from face images: Human vs.
  machine performance}.
\newblock \bibinfo{journal}{\emph{IEEE transactions on pattern analysis and
  machine intelligence}} \bibinfo{volume}{37}, \bibinfo{number}{6}
  (\bibinfo{year}{2014}), \bibinfo{pages}{1148--1161}.
\newblock


\bibitem[\protect\citeauthoryear{Han, Wang, Chen, Chen, Guo, Liu, Tang, Xiao,
  Xu, Xu, et~al\mbox{.}}{Han et~al\mbox{.}}{2022}]%
        {han2022survey}
\bibfield{author}{\bibinfo{person}{Kai Han}, \bibinfo{person}{Yunhe Wang},
  \bibinfo{person}{Hanting Chen}, \bibinfo{person}{Xinghao Chen},
  \bibinfo{person}{Jianyuan Guo}, \bibinfo{person}{Zhenhua Liu},
  \bibinfo{person}{Yehui Tang}, \bibinfo{person}{An Xiao},
  \bibinfo{person}{Chunjing Xu}, \bibinfo{person}{Yixing Xu}, {et~al\mbox{.}}}
  \bibinfo{year}{2022}\natexlab{}.
\newblock \showarticletitle{A survey on vision transformer}.
\newblock \bibinfo{journal}{\emph{IEEE transactions on pattern analysis and
  machine intelligence}} \bibinfo{volume}{45}, \bibinfo{number}{1}
  (\bibinfo{year}{2022}), \bibinfo{pages}{87--110}.
\newblock


\bibitem[\protect\citeauthoryear{Hassani and Khasahmadi}{Hassani and
  Khasahmadi}{2020}]%
        {hassani2020contrastive}
\bibfield{author}{\bibinfo{person}{Kaveh Hassani} {and}
  \bibinfo{person}{Amir~Hosein Khasahmadi}.} \bibinfo{year}{2020}\natexlab{}.
\newblock \showarticletitle{Contrastive multi-view representation learning on
  graphs}. In \bibinfo{booktitle}{\emph{International Conference on Machine
  Learning}}. PMLR, \bibinfo{pages}{4116--4126}.
\newblock


\bibitem[\protect\citeauthoryear{Hiba and Keller}{Hiba and Keller}{2021}]%
        {hiba2021hierarchical}
\bibfield{author}{\bibinfo{person}{Shakediel Hiba} {and} \bibinfo{person}{Yosi
  Keller}.} \bibinfo{year}{2021}\natexlab{}.
\newblock \showarticletitle{Hierarchical attention-based age estimation and
  Bias estimation}.
\newblock \bibinfo{journal}{\emph{arXiv preprint arXiv:2103.09882}}
  (\bibinfo{year}{2021}).
\newblock


\bibitem[\protect\citeauthoryear{Hu, Wen, Wang, Wang, Hong, and Yan}{Hu
  et~al\mbox{.}}{2016}]%
        {hu2016facial}
\bibfield{author}{\bibinfo{person}{Zhenzhen Hu}, \bibinfo{person}{Yonggang
  Wen}, \bibinfo{person}{Jianfeng Wang}, \bibinfo{person}{Meng Wang},
  \bibinfo{person}{Richang Hong}, {and} \bibinfo{person}{Shuicheng Yan}.}
  \bibinfo{year}{2016}\natexlab{}.
\newblock \showarticletitle{Facial age estimation with age difference}.
\newblock \bibinfo{journal}{\emph{IEEE Transactions on Image Processing}}
  \bibinfo{volume}{26}, \bibinfo{number}{7} (\bibinfo{year}{2016}),
  \bibinfo{pages}{3087--3097}.
\newblock


\bibitem[\protect\citeauthoryear{Huang, Liu, Van~der Maaten, and
  Weinberger}{Huang et~al\mbox{.}}{2018}]%
        {huang2018condensenet}
\bibfield{author}{\bibinfo{person}{Gao Huang}, \bibinfo{person}{Shichen Liu},
  \bibinfo{person}{Laurens Van~der Maaten}, {and} \bibinfo{person}{Kilian~Q
  Weinberger}.} \bibinfo{year}{2018}\natexlab{}.
\newblock \showarticletitle{Condensenet: An efficient densenet using learned
  group convolutions}. In \bibinfo{booktitle}{\emph{Proceedings of the IEEE
  conference on computer vision and pattern recognition}}.
  \bibinfo{pages}{2752--2761}.
\newblock


\bibitem[\protect\citeauthoryear{Kipf and Welling}{Kipf and Welling}{[n. d.]}]%
        {kipfsemi}
\bibfield{author}{\bibinfo{person}{Thomas~N Kipf} {and} \bibinfo{person}{Max
  Welling}.} \bibinfo{year}{[n. d.]}\natexlab{}.
\newblock \showarticletitle{Semi-Supervised Classification with Graph
  Convolutional Networks}. In \bibinfo{booktitle}{\emph{International
  Conference on Learning Representations}}.
\newblock


\bibitem[\protect\citeauthoryear{Landrieu and Simonovsky}{Landrieu and
  Simonovsky}{2018}]%
        {landrieu2018large}
\bibfield{author}{\bibinfo{person}{Loic Landrieu} {and} \bibinfo{person}{Martin
  Simonovsky}.} \bibinfo{year}{2018}\natexlab{}.
\newblock \showarticletitle{Large-scale point cloud semantic segmentation with
  superpoint graphs}. In \bibinfo{booktitle}{\emph{Proceedings of the IEEE
  Conference on Computer Vision and Pattern Recognition}}.
  \bibinfo{pages}{4558--4567}.
\newblock


\bibitem[\protect\citeauthoryear{Levi and Hassner}{Levi and Hassner}{2015}]%
        {levi2015age}
\bibfield{author}{\bibinfo{person}{Gil Levi} {and} \bibinfo{person}{Tal
  Hassner}.} \bibinfo{year}{2015}\natexlab{}.
\newblock \showarticletitle{Age and gender classification using convolutional
  neural networks}. In \bibinfo{booktitle}{\emph{Proceedings of the IEEE
  Conference on Computer Vision and Pattern Recognition Workshops}}.
  \bibinfo{pages}{34--42}.
\newblock


\bibitem[\protect\citeauthoryear{Li, Lu, Feng, Xu, Zhou, and Tian}{Li
  et~al\mbox{.}}{2019}]%
        {li2019bridgenet}
\bibfield{author}{\bibinfo{person}{Wanhua Li}, \bibinfo{person}{Jiwen Lu},
  \bibinfo{person}{Jianjiang Feng}, \bibinfo{person}{Chunjing Xu},
  \bibinfo{person}{Jie Zhou}, {and} \bibinfo{person}{Qi Tian}.}
  \bibinfo{year}{2019}\natexlab{}.
\newblock \showarticletitle{Bridgenet: A continuity-aware probabilistic network
  for age estimation}. In \bibinfo{booktitle}{\emph{Proceedings of the IEEE/CVF
  Conference on Computer Vision and Pattern Recognition}}.
  \bibinfo{pages}{1145--1154}.
\newblock


\bibitem[\protect\citeauthoryear{Lin, Maire, Belongie, Hays, Perona, Ramanan,
  Doll{\'a}r, and Zitnick}{Lin et~al\mbox{.}}{2014}]%
        {lin2014microsoft}
\bibfield{author}{\bibinfo{person}{Tsung-Yi Lin}, \bibinfo{person}{Michael
  Maire}, \bibinfo{person}{Serge Belongie}, \bibinfo{person}{James Hays},
  \bibinfo{person}{Pietro Perona}, \bibinfo{person}{Deva Ramanan},
  \bibinfo{person}{Piotr Doll{\'a}r}, {and} \bibinfo{person}{C~Lawrence
  Zitnick}.} \bibinfo{year}{2014}\natexlab{}.
\newblock \showarticletitle{Microsoft coco: Common objects in context}. In
  \bibinfo{booktitle}{\emph{Computer Vision--ECCV 2014: 13th European
  Conference, Zurich, Switzerland, September 6-12, 2014, Proceedings, Part V
  13}}. Springer, \bibinfo{pages}{740--755}.
\newblock


\bibitem[\protect\citeauthoryear{Liu, Sun, Zhang, Wu, Yu, and Sun}{Liu
  et~al\mbox{.}}{2020}]%
        {liu2020similarity}
\bibfield{author}{\bibinfo{person}{Hao Liu}, \bibinfo{person}{Penghui Sun},
  \bibinfo{person}{Jiaqiang Zhang}, \bibinfo{person}{Suping Wu},
  \bibinfo{person}{Zhenhua Yu}, {and} \bibinfo{person}{Xuehong Sun}.}
  \bibinfo{year}{2020}\natexlab{}.
\newblock \showarticletitle{Similarity-aware and variational deep adversarial
  learning for robust facial age estimation}.
\newblock \bibinfo{journal}{\emph{IEEE Transactions on Multimedia}}
  \bibinfo{volume}{22}, \bibinfo{number}{7} (\bibinfo{year}{2020}),
  \bibinfo{pages}{1808--1822}.
\newblock


\bibitem[\protect\citeauthoryear{Liu, Li, Kan, Zhang, Wu, Liu, Han, Shan, and
  Chen}{Liu et~al\mbox{.}}{2015}]%
        {liu2015agenet}
\bibfield{author}{\bibinfo{person}{Xin Liu}, \bibinfo{person}{Shaoxin Li},
  \bibinfo{person}{Meina Kan}, \bibinfo{person}{Jie Zhang},
  \bibinfo{person}{Shuzhe Wu}, \bibinfo{person}{Wenxian Liu},
  \bibinfo{person}{Hu Han}, \bibinfo{person}{Shiguang Shan}, {and}
  \bibinfo{person}{Xilin Chen}.} \bibinfo{year}{2015}\natexlab{}.
\newblock \showarticletitle{Agenet: Deeply learned regressor and classifier for
  robust apparent age estimation}. In \bibinfo{booktitle}{\emph{Proceedings of
  the IEEE International Conference on Computer Vision Workshops}}.
  \bibinfo{pages}{16--24}.
\newblock


\bibitem[\protect\citeauthoryear{Liu, Lin, Cao, Hu, Wei, Zhang, Lin, and
  Guo}{Liu et~al\mbox{.}}{2021}]%
        {liu2021swin}
\bibfield{author}{\bibinfo{person}{Ze Liu}, \bibinfo{person}{Yutong Lin},
  \bibinfo{person}{Yue Cao}, \bibinfo{person}{Han Hu}, \bibinfo{person}{Yixuan
  Wei}, \bibinfo{person}{Zheng Zhang}, \bibinfo{person}{Stephen Lin}, {and}
  \bibinfo{person}{Baining Guo}.} \bibinfo{year}{2021}\natexlab{}.
\newblock \showarticletitle{Swin transformer: Hierarchical vision transformer
  using shifted windows}. In \bibinfo{booktitle}{\emph{Proceedings of the
  IEEE/CVF International Conference on Computer Vision}}.
  \bibinfo{pages}{10012--10022}.
\newblock


\bibitem[\protect\citeauthoryear{Liu, Mao, Wu, Feichtenhofer, Darrell, and
  Xie}{Liu et~al\mbox{.}}{2022}]%
        {liu2022convnet}
\bibfield{author}{\bibinfo{person}{Zhuang Liu}, \bibinfo{person}{Hanzi Mao},
  \bibinfo{person}{Chao-Yuan Wu}, \bibinfo{person}{Christoph Feichtenhofer},
  \bibinfo{person}{Trevor Darrell}, {and} \bibinfo{person}{Saining Xie}.}
  \bibinfo{year}{2022}\natexlab{}.
\newblock \showarticletitle{A convnet for the 2020s}. In
  \bibinfo{booktitle}{\emph{Proceedings of the IEEE/CVF Conference on Computer
  Vision and Pattern Recognition}}. \bibinfo{pages}{11976--11986}.
\newblock


\bibitem[\protect\citeauthoryear{Luu, Seshadri, Savvides, Bui, and Suen}{Luu
  et~al\mbox{.}}{2011}]%
        {luu2011contourlet}
\bibfield{author}{\bibinfo{person}{Khoa Luu}, \bibinfo{person}{Keshav
  Seshadri}, \bibinfo{person}{Marios Savvides}, \bibinfo{person}{Tien~D Bui},
  {and} \bibinfo{person}{Ching~Y Suen}.} \bibinfo{year}{2011}\natexlab{}.
\newblock \showarticletitle{Contourlet appearance model for facial age
  estimation}. In \bibinfo{booktitle}{\emph{2011 international joint conference
  on biometrics (IJCB)}}. IEEE, \bibinfo{pages}{1--8}.
\newblock


\bibitem[\protect\citeauthoryear{Mao, Qi, Chen, Li, Duan, Ye, He, and Xue}{Mao
  et~al\mbox{.}}{2022}]%
        {mao2022towards}
\bibfield{author}{\bibinfo{person}{Xiaofeng Mao}, \bibinfo{person}{Gege Qi},
  \bibinfo{person}{Yuefeng Chen}, \bibinfo{person}{Xiaodan Li},
  \bibinfo{person}{Ranjie Duan}, \bibinfo{person}{Shaokai Ye},
  \bibinfo{person}{Yuan He}, {and} \bibinfo{person}{Hui Xue}.}
  \bibinfo{year}{2022}\natexlab{}.
\newblock \showarticletitle{Towards robust vision transformer}. In
  \bibinfo{booktitle}{\emph{Proceedings of the IEEE/CVF Conference on Computer
  Vision and Pattern Recognition}}. \bibinfo{pages}{12042--12051}.
\newblock


\bibitem[\protect\citeauthoryear{Mo, Peng, Xu, Shi, and Zhu}{Mo
  et~al\mbox{.}}{2022}]%
        {mo2022simple}
\bibfield{author}{\bibinfo{person}{Yujie Mo}, \bibinfo{person}{Liang Peng},
  \bibinfo{person}{Jie Xu}, \bibinfo{person}{Xiaoshuang Shi}, {and}
  \bibinfo{person}{Xiaofeng Zhu}.} \bibinfo{year}{2022}\natexlab{}.
\newblock \showarticletitle{Simple unsupervised graph representation learning}.
  In \bibinfo{booktitle}{\emph{Proceedings of the AAAI Conference on Artificial
  Intelligence}}, Vol.~\bibinfo{volume}{36}. \bibinfo{pages}{7797--7805}.
\newblock


\bibitem[\protect\citeauthoryear{Niu, Zhou, Wang, Gao, and Hua}{Niu
  et~al\mbox{.}}{2016}]%
        {niu2016ordinal}
\bibfield{author}{\bibinfo{person}{Zhenxing Niu}, \bibinfo{person}{Mo Zhou},
  \bibinfo{person}{Le Wang}, \bibinfo{person}{Xinbo Gao}, {and}
  \bibinfo{person}{Gang Hua}.} \bibinfo{year}{2016}\natexlab{}.
\newblock \showarticletitle{Ordinal regression with multiple output cnn for age
  estimation}. In \bibinfo{booktitle}{\emph{Proceedings of the IEEE conference
  on computer vision and pattern recognition}}. \bibinfo{pages}{4920--4928}.
\newblock


\bibitem[\protect\citeauthoryear{Rothe, Timofte, and Van~Gool}{Rothe
  et~al\mbox{.}}{2016}]%
        {rothe2016some}
\bibfield{author}{\bibinfo{person}{Rasmus Rothe}, \bibinfo{person}{Radu
  Timofte}, {and} \bibinfo{person}{Luc Van~Gool}.}
  \bibinfo{year}{2016}\natexlab{}.
\newblock \showarticletitle{Some like it hot-visual guidance for preference
  prediction}. In \bibinfo{booktitle}{\emph{Proceedings of the IEEE conference
  on computer vision and pattern recognition}}. \bibinfo{pages}{5553--5561}.
\newblock


\bibitem[\protect\citeauthoryear{Rothe, Timofte, and Van~Gool}{Rothe
  et~al\mbox{.}}{2018}]%
        {rothe2018deep}
\bibfield{author}{\bibinfo{person}{Rasmus Rothe}, \bibinfo{person}{Radu
  Timofte}, {and} \bibinfo{person}{Luc Van~Gool}.}
  \bibinfo{year}{2018}\natexlab{}.
\newblock \showarticletitle{Deep expectation of real and apparent age from a
  single image without facial landmarks}.
\newblock \bibinfo{journal}{\emph{International Journal of Computer Vision}}
  \bibinfo{volume}{126}, \bibinfo{number}{2-4} (\bibinfo{year}{2018}),
  \bibinfo{pages}{144--157}.
\newblock


\bibitem[\protect\citeauthoryear{Shen, Guo, Wang, Zhao, Wang, and Yuille}{Shen
  et~al\mbox{.}}{2019}]%
        {shen2019deep}
\bibfield{author}{\bibinfo{person}{Wei Shen}, \bibinfo{person}{Yilu Guo},
  \bibinfo{person}{Yan Wang}, \bibinfo{person}{Kai Zhao}, \bibinfo{person}{Bo
  Wang}, {and} \bibinfo{person}{Alan Yuille}.} \bibinfo{year}{2019}\natexlab{}.
\newblock \showarticletitle{Deep differentiable random forests for age
  estimation}.
\newblock \bibinfo{journal}{\emph{IEEE transactions on pattern analysis and
  machine intelligence}} \bibinfo{volume}{43}, \bibinfo{number}{2}
  (\bibinfo{year}{2019}), \bibinfo{pages}{404--419}.
\newblock


\bibitem[\protect\citeauthoryear{Shen, Guo, Wang, Zhao, Wang, and Yuille}{Shen
  et~al\mbox{.}}{2018}]%
        {shen2018deep}
\bibfield{author}{\bibinfo{person}{Wei Shen}, \bibinfo{person}{Yilu Guo},
  \bibinfo{person}{Yan Wang}, \bibinfo{person}{Kai Zhao}, \bibinfo{person}{Bo
  Wang}, {and} \bibinfo{person}{Alan~L Yuille}.}
  \bibinfo{year}{2018}\natexlab{}.
\newblock \showarticletitle{Deep regression forests for age estimation}. In
  \bibinfo{booktitle}{\emph{Proceedings of the IEEE Conference on Computer
  Vision and Pattern Recognition}}. \bibinfo{pages}{2304--2313}.
\newblock


\bibitem[\protect\citeauthoryear{Shi, Yan, Li, Li, Liu, Sun, Wang, Zhang, Zou,
  and Wu}{Shi et~al\mbox{.}}{2020}]%
        {shi2020fetal}
\bibfield{author}{\bibinfo{person}{Wen Shi}, \bibinfo{person}{Guohui Yan},
  \bibinfo{person}{Yamin Li}, \bibinfo{person}{Haotian Li},
  \bibinfo{person}{Tingting Liu}, \bibinfo{person}{Cong Sun},
  \bibinfo{person}{Guangbin Wang}, \bibinfo{person}{Yi Zhang},
  \bibinfo{person}{Yu Zou}, {and} \bibinfo{person}{Dan Wu}.}
  \bibinfo{year}{2020}\natexlab{}.
\newblock \showarticletitle{Fetal brain age estimation and anomaly detection
  using attention-based deep ensembles with uncertainty}.
\newblock \bibinfo{journal}{\emph{NeuroImage}}  \bibinfo{volume}{223}
  (\bibinfo{year}{2020}), \bibinfo{pages}{117316}.
\newblock


\bibitem[\protect\citeauthoryear{Shin, Lee, and Kim}{Shin
  et~al\mbox{.}}{2022}]%
        {shin2022moving}
\bibfield{author}{\bibinfo{person}{Nyeong-Ho Shin}, \bibinfo{person}{Seon-Ho
  Lee}, {and} \bibinfo{person}{Chang-Su Kim}.} \bibinfo{year}{2022}\natexlab{}.
\newblock \showarticletitle{Moving window regression: a novel approach to
  ordinal regression}. In \bibinfo{booktitle}{\emph{Proceedings of the IEEE/CVF
  Conference on Computer Vision and Pattern Recognition}}.
  \bibinfo{publisher}{IEEE}, \bibinfo{pages}{18760--18769}.
\newblock


\bibitem[\protect\citeauthoryear{Shoba and Sam}{Shoba and Sam}{2022}]%
        {shoba2022adaptive}
\bibfield{author}{\bibinfo{person}{V~Betcy~Thanga Shoba} {and}
  \bibinfo{person}{I~Shatheesh Sam}.} \bibinfo{year}{2022}\natexlab{}.
\newblock \showarticletitle{Adaptive deep feature learning based Softmax
  regressive classification for aging facial recognition}.
\newblock \bibinfo{journal}{\emph{Multimedia Tools and Applications}}
  (\bibinfo{year}{2022}), \bibinfo{pages}{1--29}.
\newblock


\bibitem[\protect\citeauthoryear{Tang, Deng, Huang, and Zhao}{Tang
  et~al\mbox{.}}{2014}]%
        {tang2014compressed}
\bibfield{author}{\bibinfo{person}{Jiexiong Tang}, \bibinfo{person}{Chenwei
  Deng}, \bibinfo{person}{Guang-Bin Huang}, {and} \bibinfo{person}{Baojun
  Zhao}.} \bibinfo{year}{2014}\natexlab{}.
\newblock \showarticletitle{Compressed-domain ship detection on spaceborne
  optical image using deep neural network and extreme learning machine}.
\newblock \bibinfo{journal}{\emph{IEEE transactions on geoscience and remote
  sensing}} \bibinfo{volume}{53}, \bibinfo{number}{3} (\bibinfo{year}{2014}),
  \bibinfo{pages}{1174--1185}.
\newblock


\bibitem[\protect\citeauthoryear{Velickovic, Fedus, Hamilton, Li{\`o}, Bengio,
  and Hjelm}{Velickovic et~al\mbox{.}}{2019}]%
        {velickovic2019deep}
\bibfield{author}{\bibinfo{person}{Petar Velickovic}, \bibinfo{person}{William
  Fedus}, \bibinfo{person}{William~L Hamilton}, \bibinfo{person}{Pietro
  Li{\`o}}, \bibinfo{person}{Yoshua Bengio}, {and} \bibinfo{person}{R~Devon
  Hjelm}.} \bibinfo{year}{2019}\natexlab{}.
\newblock \showarticletitle{Deep graph infomax.}
\newblock \bibinfo{journal}{\emph{ICLR (Poster)}} \bibinfo{volume}{2},
  \bibinfo{number}{3} (\bibinfo{year}{2019}), \bibinfo{pages}{4}.
\newblock


\bibitem[\protect\citeauthoryear{Wang, Sanchez, and Li}{Wang
  et~al\mbox{.}}{2022}]%
        {wang2022improving}
\bibfield{author}{\bibinfo{person}{Haoyi Wang}, \bibinfo{person}{Victor
  Sanchez}, {and} \bibinfo{person}{Chang-Tsun Li}.}
  \bibinfo{year}{2022}\natexlab{}.
\newblock \showarticletitle{Improving face-based age estimation with
  attention-based dynamic patch fusion}.
\newblock \bibinfo{journal}{\emph{IEEE Transactions on Image Processing}}
  \bibinfo{volume}{31} (\bibinfo{year}{2022}), \bibinfo{pages}{1084--1096}.
\newblock


\bibitem[\protect\citeauthoryear{Wang, Guo, and Kambhamettu}{Wang
  et~al\mbox{.}}{2015}]%
        {wang2015deeply}
\bibfield{author}{\bibinfo{person}{Xiaolong Wang}, \bibinfo{person}{Rui Guo},
  {and} \bibinfo{person}{Chandra Kambhamettu}.}
  \bibinfo{year}{2015}\natexlab{}.
\newblock \showarticletitle{Deeply-learned feature for age estimation}. In
  \bibinfo{booktitle}{\emph{2015 IEEE Winter Conference on Applications of
  Computer Vision}}. IEEE, \bibinfo{pages}{534--541}.
\newblock


\bibitem[\protect\citeauthoryear{Xie, Zhang, Yan, and Liu}{Xie
  et~al\mbox{.}}{2015}]%
        {xie2015hybrid}
\bibfield{author}{\bibinfo{person}{Guo-Sen Xie}, \bibinfo{person}{Xu-Yao
  Zhang}, \bibinfo{person}{Shuicheng Yan}, {and} \bibinfo{person}{Cheng-Lin
  Liu}.} \bibinfo{year}{2015}\natexlab{}.
\newblock \showarticletitle{Hybrid CNN and dictionary-based models for scene
  recognition and domain adaptation}.
\newblock \bibinfo{journal}{\emph{IEEE Transactions on Circuits and Systems for
  Video Technology}} \bibinfo{volume}{27}, \bibinfo{number}{6}
  (\bibinfo{year}{2015}), \bibinfo{pages}{1263--1274}.
\newblock


\bibitem[\protect\citeauthoryear{Yan, Xiong, and Lin}{Yan
  et~al\mbox{.}}{2018}]%
        {yan2018spatial}
\bibfield{author}{\bibinfo{person}{Sijie Yan}, \bibinfo{person}{Yuanjun Xiong},
  {and} \bibinfo{person}{Dahua Lin}.} \bibinfo{year}{2018}\natexlab{}.
\newblock \showarticletitle{Spatial temporal graph convolutional networks for
  skeleton-based action recognition}. In \bibinfo{booktitle}{\emph{Proceedings
  of the AAAI conference on artificial intelligence}},
  Vol.~\bibinfo{volume}{32}.
\newblock


\bibitem[\protect\citeauthoryear{Yue, Yongbin, Ziwei, Sarma, and Bronstein}{Yue
  et~al\mbox{.}}{2019}]%
        {yue2019dynamic}
\bibfield{author}{\bibinfo{person}{Wang Yue}, \bibinfo{person}{Sun Yongbin},
  \bibinfo{person}{Liu Ziwei}, \bibinfo{person}{Sanjay~E Sarma}, {and}
  \bibinfo{person}{Michael~M Bronstein}.} \bibinfo{year}{2019}\natexlab{}.
\newblock \showarticletitle{Dynamic graph cnn for learning on point clouds}.
\newblock \bibinfo{journal}{\emph{ACM Transactions on Graphics (TOG)}}
  \bibinfo{volume}{38}, \bibinfo{number}{5} (\bibinfo{year}{2019}).
\newblock


\bibitem[\protect\citeauthoryear{Zeng and Xie}{Zeng and Xie}{2021}]%
        {zeng2021contrastive}
\bibfield{author}{\bibinfo{person}{Jiaqi Zeng} {and} \bibinfo{person}{Pengtao
  Xie}.} \bibinfo{year}{2021}\natexlab{}.
\newblock \showarticletitle{Contrastive self-supervised learning for graph
  classification}. In \bibinfo{booktitle}{\emph{Proceedings of the AAAI
  Conference on Artificial Intelligence}}, Vol.~\bibinfo{volume}{35}.
  \bibinfo{pages}{10824--10832}.
\newblock


\bibitem[\protect\citeauthoryear{Zhang, Liu, Xu, and Zhu}{Zhang
  et~al\mbox{.}}{2019a}]%
        {zhang2019c3ae}
\bibfield{author}{\bibinfo{person}{Chao Zhang}, \bibinfo{person}{Shuaicheng
  Liu}, \bibinfo{person}{Xun Xu}, {and} \bibinfo{person}{Ce Zhu}.}
  \bibinfo{year}{2019}\natexlab{a}.
\newblock \showarticletitle{C3AE: Exploring the limits of compact model for age
  estimation}. In \bibinfo{booktitle}{\emph{Proceedings of the IEEE/CVF
  Conference on Computer Vision and Pattern Recognition}}.
  \bibinfo{publisher}{IEEE}, \bibinfo{pages}{12587--12596}.
\newblock


\bibitem[\protect\citeauthoryear{Zhang, Liu, Yuan, Guo, Gao, Zhao, and
  Ma}{Zhang et~al\mbox{.}}{2019b}]%
        {zhang2019fine}
\bibfield{author}{\bibinfo{person}{Ke Zhang}, \bibinfo{person}{Na Liu},
  \bibinfo{person}{Xingfang Yuan}, \bibinfo{person}{Xinyao Guo},
  \bibinfo{person}{Ce Gao}, \bibinfo{person}{Zhenbing Zhao}, {and}
  \bibinfo{person}{Zhanyu Ma}.} \bibinfo{year}{2019}\natexlab{b}.
\newblock \showarticletitle{Fine-grained age estimation in the wild with
  attention LSTM networks}.
\newblock \bibinfo{journal}{\emph{IEEE Transactions on Circuits and Systems for
  Video Technology}} \bibinfo{volume}{30}, \bibinfo{number}{9}
  (\bibinfo{year}{2019}), \bibinfo{pages}{3140--3152}.
\newblock


\bibitem[\protect\citeauthoryear{Zhang, Liu, Li, et~al\mbox{.}}{Zhang
  et~al\mbox{.}}{2017}]%
        {zhang2017quantifying}
\bibfield{author}{\bibinfo{person}{Yunxuan Zhang}, \bibinfo{person}{Li Liu},
  \bibinfo{person}{Cheng Li}, {et~al\mbox{.}}} \bibinfo{year}{2017}\natexlab{}.
\newblock \showarticletitle{Quantifying facial age by posterior of age
  comparisons}.
\newblock \bibinfo{journal}{\emph{arXiv preprint arXiv:1708.09687}}
  (\bibinfo{year}{2017}).
\newblock


\bibitem[\protect\citeauthoryear{Zhang and Yeung}{Zhang and Yeung}{2010}]%
        {zhang2010multi}
\bibfield{author}{\bibinfo{person}{Yu Zhang} {and} \bibinfo{person}{Dit-Yan
  Yeung}.} \bibinfo{year}{2010}\natexlab{}.
\newblock \showarticletitle{Multi-task warped gaussian process for personalized
  age estimation}. In \bibinfo{booktitle}{\emph{2010 IEEE computer society
  conference on computer vision and pattern recognition}}. IEEE,
  \bibinfo{pages}{2622--2629}.
\newblock


\bibitem[\protect\citeauthoryear{Zhu, Xu, Yu, Liu, Wu, and Wang}{Zhu
  et~al\mbox{.}}{2021}]%
        {zhu2021graph}
\bibfield{author}{\bibinfo{person}{Yanqiao Zhu}, \bibinfo{person}{Yichen Xu},
  \bibinfo{person}{Feng Yu}, \bibinfo{person}{Qiang Liu}, \bibinfo{person}{Shu
  Wu}, {and} \bibinfo{person}{Liang Wang}.} \bibinfo{year}{2021}\natexlab{}.
\newblock \showarticletitle{Graph contrastive learning with adaptive
  augmentation}. In \bibinfo{booktitle}{\emph{Proceedings of the Web Conference
  2021}}. \bibinfo{pages}{2069--2080}.
\newblock


\end{thebibliography}
	
\end{document}